\documentclass[review]{elsarticle}

\usepackage{hyperref}
\usepackage{times}
\usepackage{epsfig}
\usepackage{graphicx}
\usepackage{amsmath}
\usepackage{amssymb}
\usepackage{algorithm}
\usepackage{algcompatible}
\usepackage{url}

\DeclareMathAlphabet{\mathcal}{OMS}{cmsy}{m}{n}

\usepackage{amsthm}
\usepackage{bm} %use command \bm{}
\usepackage{amsfonts}
\usepackage{latexsym}
\usepackage{multirow}
\usepackage{subfigure}
\usepackage[T1]{fontenc} % Encodage ISO-8859-1.
\usepackage{graphicx} %Pour les figures, disponible sur : ctan
\usepackage{mathptmx} %Times Font for latin and Symbol font for greek.
\usepackage{subfigure}
\usepackage{fancyhdr}
\usepackage{float}
\usepackage{color}
\usepackage{hhline}
\usepackage{epstopdf}
\usepackage{ctable}
\usepackage[percent]{overpic}

\usepackage{rotating}

\newcommand{\x}{\textbf{x}}
\newcommand{\y}{\textbf{y}}

\newcommand{\X}{\textbf{X}}

\usepackage{calligra}
\DeclareMathAlphabet{\mathcalligra}{T1}{calligra}{m}{n}
\newcommand{\STAB}[1]{\begin{tabular}{@{}c@{}}#1\end{tabular}}

\definecolor{hotpink}{rgb}{1.0, 0.41, 0.71}

\newcommand{\cblue}{\textcolor{black}}

\newcommand{\Real}{\mathbb{R}}

\newcommand{\btheta}{\bm{\theta}}
\newcommand{\thetab}{\bm{\theta}}
 
\newcommand{\bSigma}{\bm{\Sigma}}
\newcommand{\bDelta}{\bm{\Delta}}

\newcommand{\bmu}{\bm \mu}

\newcommand{\bb}{{\bf b}}
\newcommand{\Y}{{\bf Y}}

\newcommand{\yb}{{\bf y}}

\newcommand{\xb}{{\bf x}}

\newcommand{\Wb}{{\bf W}}

\newcommand{\eR}{{\mathbb{R}}}

\newcommand{\Rc}{{\mathcal{R}}}
\newcommand{\Nc}{{\mathcal{N}}}

\newcommand{\Bc}{{\mathrm{Ber}}}
\newcommand{\Mc}{{\mathrm{Mult}}}

\algnewcommand\INITIALIZATION{\item[\textbf{Initialization.}]}%
\algnewcommand\ITERATION{\item[\textbf{Iterative steps.}]}%

\journal{XXX}

\begin{document}

\begin{frontmatter}

\title{Efficient Bayes Inference in Neural Networks through Adaptive Importance Sampling}

\author[mymainaddress]{Yunshi Huang}

\author[mysecondaryaddress]{Emilie Chouzenoux\corref{mycorrespondingauthor}}
\cortext[mycorrespondingauthor]{Corresponding author}
\ead{emilie.chouzenoux@centralesupelec.fr}

\author[mythirdaddress]{Víctor Elvira}

\author[mysecondaryaddress]{Jean-Christophe Pesquet}
\address[mymainaddress]{ETS Montr\'eal, Canada}
\address[mysecondaryaddress]{CVN, Inria Saclay, CentraleSupélec, Universit\'e Paris-Saclay, France}
\address[mythirdaddress]{University of Edinburgh, UK}

\begin{abstract}
Bayesian neural networks (BNNs) have received an increased interest in the last years. In BNNs, a complete posterior distribution of the unknown weight and bias parameters of the network is produced during the training stage. This probabilistic estimation offers several advantages with respect to point-wise estimates, in particular, the ability to provide uncertainty quantification when predicting new data. This feature inherent to the Bayesian paradigm, is useful in countless machine learning applications. It is particularly appealing in areas where decision-making has a crucial impact, such as medical healthcare or autonomous driving. The main challenge of BNNs is the computational cost of the training procedure since Bayesian techniques often face a severe curse of dimensionality. Adaptive importance sampling (AIS) is one of the most prominent Monte Carlo methodologies benefiting from sounded convergence guarantees and ease for adaptation. This work aims to show that AIS constitutes a successful approach for designing BNNs. More precisely, we propose a novel algorithm named PMCnet that includes an efficient adaptation mechanism, exploiting geometric information on the complex (often multimodal) posterior distribution. % without increasing the computational cost. 
Numerical results illustrate the excellent performance and the improved exploration capabilities of the proposed method for both shallow and deep neural networks.
\end{abstract}


\end{frontmatter}

%\linenumbers

\section{Introduction}
\label{sec:intro}
Deep neural networks (DNNs) are often the current state-of-the-art for solving a wide range of diverse tasks in
machine learning. They consist in a cascade of linear and nonlinear operators that are usually optimized from large amounts of labeled data using back-propagation techniques. 
%through supervision by utilizing large amounts of labeled data. DNN parameters are typically optimized by back-propagation algorithm. 
%stochastic gradient descent, through the 
However, this optimization procedure often relies on ad-hoc machinery which may not lead to relevant local minima without good numerical recipes. Furthermore, it provides no information regarding the uncertainty of the obtained predictions. However, uncertainty is inherent in machine learning, stemming either from \cblue{the noise in the data values, the statistical variability of the data distribution,} the sample selection procedure, and the imperfect nature of any developed model. Quantifying this uncertainty is of paramount importance in a wide array of applied fields such as self-driving cars, medicine, or forecasting. Bayesian neural network (BNN) approaches offer a grounded theoretical framework to tackle model uncertainty in the context of DNNs~\cite{BishopBNN}.

In the Bayesian inference framework, a statistical model is assumed between the unknown parameters and the given data in order to build a posterior distribution of those unknowns conditioned to the data. However, for most practical models, the posterior distribution is not available in a closed form, mostly due to intractable integrals, and approximations must be performed via Monte Carlo (MC) methods \cite{Robert04}. Importance sampling (IS) is a Monte Carlo family of methods that consists in simulating random samples from a proposal distribution and weighting them properly with the aim of building consistent estimators of the moments of the posterior distribution. The performance of IS depends on the choice of the proposal distribution \cite{elvira2015efficient,elvira2016heretical,elvira2019generalized}. Adaptive IS (AIS) is an iterative version of IS where the proposal distributions are adapted based on their performance at previous iterations \cite{bugallo2017adaptive}. In the last decade, many AIS algorithms have been proposed in the literature \cite{Cappe08,CORNUET12,APIS14,APIS15,martino2015layered,elvira2017improving,el2018robust}. However, two main challenges still exist and need to be tackled. First, 
%although %all the algorithms adapt the location parameter (i.e., the mean) of the proposal, only f
few AIS algorithms adapt the scale parameter, which is problematic when the unknowns have different orders of magnitude. 
For instance, the covariance matrix is adapted via robust moment matching strategies in \cite{el2018robust,ellaham2019recursive}. Second, the use of the geometry of the target for adaptation rule has only been explored scarcely in the recent AIS literature \cite{schuster2015gradient,fasiolo2018langevin,elvira2015gradient}. On the one hand, optimization-based schemes have been proposed to accelerate MCMC algorithms convergence~\cite{roberts2002langevin,durmus2016efficient,Schreck16}, such as in Metropolis adjusted Langevin algorithm (MALA), which combines an unadjusted Langevin (ULA) update with an acceptance-rejection step. MALA performance can be further improved by a preconditioning strategy~\cite{Fisher_MCMC_1,marnissi2018}. 
The recent SL-PMC algorithm \cite{elvira2019langevin} is up to our knowledge the only AIS-based method that exploits first and second-order information on the target to adapt both the location and scale parameters of the proposals.
%The MALA algorithm has been refined by a local adaptation of the drift term via a preconditioning or scaling matrix used in the ULA stage \cite{Fisher_MCMC_1,marnissi2018}.
% is up to our knowledge the only AIS-based algorithm that exploits the use of first and second-order information on the target to adapt both the location and scale parameters of the proposals. %The construction of this scaling matrix is a subject of research, for instance using second-order information, the Fisher metric \cite{Fisher_MCMC_1}, or majorization-minimization mechanisms \cite{marnissi2018}. The SL-PMC algorithm \cite{elvira2019langevin} is up to our knowledge the only AIS-based algorithm that exploits the use of second-order information for adapting both the location and scale parameters of the proposals. \victor{EC, please complete.}

BNN inference is usually performed using the variational Bayesian technique \cite{Honton93bnn,Barber98,Sun2019bnn}, which consists in constructing a tractable approximation to the posterior distribution (e.g., based on a mean field approximation). However, the results may be sensitive to the approximation error and to initialization. 
%few guarantees are available for those techniques whose
Promising results have recently been reached by using MC sampling strategies instead. %Those last years, there has been several works showing that promising results in terms of inference accuracy and computational cost can be reached using Monte-Carlo sampling strategies instead.
%Due to the large dimension of the problem and the multimodality of the posterior
Again, a key ingredient for good performance lies in an efficient adaptation strategy, usually by relying on tools from optimization.
The stochastic {gradient} Langevin dynamics method from \cite{WellingSGLD}, a mini-batched version of ULA, seems now to be able to reach state-of-the-art results with reasonable computational cost, as illustrated in \cite{Wang2018bnn,Nado2018}. One can also mention the Hamiltonian MC sampler with local scale adaptation, proposed in \cite{Springenberg2016bnn}. %, associated with an appropriate scale adaptation method
In {\cite{Gal2016}, the dropout in the neural network is given by an approximation to the probabilistic deep Gaussian process. In \cite{Arnaud2022}, the method called Sequential Anchored Ensembles,   trains the ensemble sequentially starting from the previous solution to reduce the computational cost of the training process.} 

{In this paper, we propose the first AIS algorithm for BNN inference. 
 IS-based methods have several advantages w.r.t. MCMC, e.g., all the generated samples are employed in the estimation (i.e., there is no ``burn-in'' period) and the corresponding adaptive schemes are more flexible (see the theoretical issues of adaptive MCMC in \cite[Section 7.6.3]{Robert04},\cite{andrieu2008tutorial}). 
 In return, the challenge is to design adaptive mechanisms for the proposal densities in order to iteratively improve the performance of the IS estimators \cite{bugallo2017adaptive}.}  
 %
%\textcolor{red}{JCP: Could we say why we expect IS to be better than MCMC ?}
% 
We develop a new strategy to adapt efficiently the proposal using a scaled ULA step. The scaling matrix is adapted via robust covariance estimators, using the weighted samples of AIS, thus avoiding the computation of a costly Hessian matrix. Another novelty is the joint mean and covariance adaptation, offering the advantage of fitting the proposal distributions locally, boosting the exploration and increasing the performance. The most noteworthy feature of the proposed novel approach is its ability to provide meaningful uncertainty quantification with a reasonable computation cost.

Numerical experiments on classification and regression problems illustrate the efficiency of our method when compared to a state-of-the-art back-propagation procedure and other BNN methods. The outline of the paper is as follows.  Section~\ref{sec:Bayes} introduces the problem and notation related to Bayesian inference in machine learning, and recall the principle of AIS with proposal adaptation. Section~\ref{sec:proposed} presents the BNN inference problem and the proposed AIS algorithm. Section \ref{sec:experiments} provides numerical results and Section \ref{sec:conclusion} concludes the paper.

\section{Motivating framework and background}
\label{sec:Bayes}
\subsection{Bayesian inference in supervised machine learning}

Supervised machine learning aims at estimating a vector of unknown parameters $\btheta \in\Real^{d_{\theta}}$ from a training set of $N_{\text{train}}$ input/output pairs of data $\left\{\x_0^{(n)},\y^{(n)}\right\}_{1\le n\le N_{\text{train}}} \in \Real^{d_{x}} \times \Real^{d_{y}}$. Let us denote by $\X_0 \in \Real^{d_x \times N_{\text{train}}}$, and $\Y \in \Real^{d_y \times N_{\text{train}}}$ the columnwise concatenation of $\left\{\x_0^{(n)}\right\}_{1\le n\le N_{\text{train}}}$, and $\left\{\y^{(n)}\right\}_{1\le n\le N_{\text{train}}}$, respectively. The unknown $\btheta$ is related to $\X_0$ and $\Y$ through a statistical model given by the likelihood function $\ell \left(\Y \left|\right. \btheta, \X_0 \right)$. The prior probabilistic knowledge about the unknown is summarized in $p\left(\btheta\right)$, $\btheta$ being assumed to be independent of $\X_0$. In probabilistic machine learning, the goal is then to infer the posterior distribution
\begin{align}
	p\left(\btheta \left|\right. \X_0, \Y \right)  
	=  \frac{\ell \left(\Y  \left|\right. \btheta, \X_0 \right)  p\left(\btheta\right)}{Z(\X_0,\Y)} & := \widetilde{\pi}\left(\btheta  \right) \nonumber\\
	& \propto \pi \left(\btheta   \right),
\end{align}
where $\pi(\btheta) := \ell \left(\Y  \left|\right. \btheta, \X_0 \right)  p\left(\btheta\right)$  and $Z = \int  \pi(\btheta) d\btheta$.\footnote{We now drop $\Y$ and $\X_0$ in $Z$, $\pi(\btheta)$, and $\widetilde \pi(\btheta)$  to alleviate the notation.}

Usually we are also interested in computing integrals of the form
%In probabilistic machine learning approaches, such as those developed for BNNs, the goal consists in characterizing the posterior distribution which is intimately related to performing a probabilistic prediction, i.e., with uncertainty quantification. This typically amounts to computing integrals taking the form
\begin{equation}
	I =  \int h(\btheta) \widetilde{\pi}(\btheta) d\btheta, % = \frac{1}{Z} \int h(\btheta) \pi(\btheta) d\btheta,
\label{eq_integral}
\end{equation}
where $h$ is any integrable function w.r.t. $\widetilde \pi(\btheta)$. However, realistic predictive models in machine learning include non-linearities (e.g., sigmoid activation functions) and loss functions corresponding to non-Gaussian potentials  (e.g., cross-entropy). Hence, neither Eq. \eqref{eq_integral} nor the normalizing constant $Z$ can be computed easily. In this case, we resort to sampling methods to find approximations to the posterior distribution and get access to the uncertainty in the estimation.

\subsection{Adaptive Importance Sampling}
%\victor{requires defining $\pi$, $f(\x)$ and the integral $I$.}
In the following, we briefly describe the basic importance sampling (IS) methodology and state-of-the-art adaptive IS (AIS) algorithms.
\subsubsection{Importance sampling (IS)}
\label{sec_IS}
Importance sampling (IS) is a Monte Carlo methodology to approximate intractable integrals. The standard IS implementation is composed of two steps. First, $K$ samples are simulated from the so-called proposal distribution $q(\cdot)$, as ${\btheta_k} \sim q(\btheta)$, $k\in \{1,\ldots,K\}$. Second, each sample is assigned an importance weight computed as
$w_k = \frac{\pi(\btheta_k)}{q(\btheta_k)}$, $k\in \{1,\ldots,K\}$. The targeted integral given by Eq. \eqref{eq_integral} can be approximated by the self-normalized IS (SNIS) estimator given by
\begin{equation}
\widetilde I = \sum_{k=1}^{K} \overline w_k h(\btheta_k),
\end{equation} 
where $\overline w_k = w_k\Big/\sum_{j=1}^K w_j$ are the normalized weights. The key lies in the selection of $q(\btheta)$, which must be nonzero for every $\btheta$ such that $h(\btheta)\widetilde\pi(\btheta)>0$. For a generic $h(\btheta)$ (or a bunch of them), a common strategy is to find the proposal $q(\btheta)$ that minimizes in some sense {(e.g., $\chi^{2}$ divergence \cite{akyildiz2021convergence})} the mismatch with the target $\widetilde \pi(\btheta)$. However, since it is usually impossible to know in advance the best proposal, adaptive mechanisms are employed. 

%Note that another IS estimator exists, namely the unnormalized IS (UIS) estimator $\hat I = \frac{1}{KZ} \sum_{k=1}^{K} w_k h(\btheta_k)$, but it can only be used when $Z$ is known. 
%
%The performance of the IS estimators strongly depends on the choice of the proposal $q(\btheta)$. 
%\cred{The proposal that minimizes the variance is given by $q(\btheta) \propto |h(\btheta)|\pi(\btheta)$ \cite{Robert04,Liu04b}. However, in all practical scenarios, it is unfeasible to use this proposal \cite[Chapter 9]{Owen13}.}

\subsubsection{Adaptive importance sampling methods (AIS)}

AIS methods are IS-based methods that improve the proposal distribution iteratively \cite{bugallo2017adaptive}. 
%\victor{AMIS/GAPIS/GIS/LAIS?}
Arguably one of the main families of AIS is the population Monte Carlo (PMC) algorithms {\cite{Cappe04,Cappe08,Elvira2021,miller2021rare}}. In PMC, the classical sampling-weighting steps are followed by an adaptation step based on resampling. %\cred{In this step, the location parameters of the proposals are moved to where previous samples are, with probability proportional to the associated weight.} 
This mechanism promotes the concentration of proposals in areas where the targeted distribution has significant probability mass \cite[Section 4.1]{elvira2017improving}. The DM-PMC in \cite{elvira2017improving} enhances the estimation and adaptation capabilities of existing PMC methods. Recently, the SL-PMC \cite{elvira2019langevin} has improved the performance of DM-PMC by incorporating geometric information on the target. 

\section{Proposed method}
\label{sec:proposed}
\subsection{Bayesian neural network model}
\label{sec:proposedmodel}
We focus on the probabilistic inference of the parameters of an $L$-layer fully connected neural network (FCNN) with $L \geq 1$, relating an input vector of dimension $d_x$ to an output vector of dimension $d_y$.
We will assume that sequences of
$N_{\text{train}}$ input entries $\{\x_0^{(n)}\}_{1\le n\le N_{\text{train}}}$ in $\Real^{d_{x}}$ and of $N_{\text{train}}$ associated output values $\left\{\y^{(n)}\right\}_{1\le n\le N_{\text{train}}}$ in $\Real^{d_{y}}$ are available.
Each layer $\ell \in \{1,\ldots,L\}$
of the considered feedforward neural model 
%between $\x_0$ and $\y$, 
is parametrized by a weight matrix $\Wb_\ell\in \eR^{S_\ell \times S_{\ell-1}}$, a bias vector $\bb_\ell\in \eR^{S_\ell}$, and  a nonlinear activation function $\Rc_{\ell}$ from $\eR^{S_\ell}$ to $\eR^{S_\ell}$. For every $n \in \left\{1,\ldots,N_{\text{train}}\right\}$
and $\ell \in \left\{1,\ldots,L\right\}$,
\begin{equation}
\x_\ell^{(n)} = \Rc_{\ell}\left(\Wb_\ell \x_{\ell-1}^{(n)} + \bb_\ell \right).
\label{eq:bnn1}
\end{equation}
%and
%\begin{equation}
%\zb^{(n)} = \Wb_{L+1} \xb_L^{(n)}.
%\label{eq:bnno}
%\end{equation}
The output vectors $\left\{\y^{(n)}\right\}_{1\le n\le N_{\text{train}}}$ are linked to $\{\xb_L^{(n)}\}_{1\le n\le N_{\text{train}}}$ through the probability density function $p(\y^{(n)} |\xb_L^{(n)})$ that depends on the machine learning task of interest. Conditionally to  $\{\xb_L^{(n)}\}_{1\le n\le N_{\text{train}}}$, $\left\{\y^{(n)}\right\}_{1\le n\le N_{\text{train}}}$ are assumed to be independent of
$\{\xb_\ell^{(n)}\}_{1\le n\le N_{\text{train}},1\le \ell\le L-1}$.
%
%for every $n \in \left\{1,\ldots,N\right\}$, $\xb_0^{(n)} \in \Real^{d_x}$ is a known input vector (typically, a vector of features), and $\vb^{(n)} \in \Real^{d_y}$ represents \cblue{i.i.d.} noise \victor{with which distribution?}. Moreover, $(\Rc_{\ell})_{1 \leq \ell \leq L}$ are known non-linear activation functions. $(\Wb_\ell,\bb_\ell)$ gathers the unknown parameters at each layer $\ell \in \left\{1,\ldots,L\right\}$ and $\Wb_{L+1}$ is an unknown linear operator acting on the output. Regarding the dimensions, we set, for every $n  \in \left\{1,\ldots,N\right\}$, and $\ell \in \left\{1,\ldots,L\right\})$, $\xb_\ell^{(n)} \in \eR^{S_\ell}$ with $S_\ell \geq 1$, so that $\bb_\ell \in \eR^{S_\ell}$, $\Wb_\ell \in \eR^{S_\ell \times S_{\ell-1}}$ and $\Rc_{\ell}: \eR^{S_\ell} \to \eR^{S_\ell}$, with $S_0 \equiv d_x$. Furthermore, for every $n  \in \left\{1,\ldots,N\right\}$, $\yb^{(n)} \in \eR^{d_y}$, so that $\Wb_{L+1} \in \eR^{d_y \times S_L}$. 
The vector
$
\btheta  = \left\{\Wb_1,\ldots,\Wb_L,\bb_1,\ldots,\bb_L\right\}
$
%\victor{be careful, in MATLAB is not stored like that.} OK !
contains all unknown parameters, with size $d_{\theta} = \sum_{\ell = 1}^L S_\ell (S_{\ell-1} + 1)$, so that we can rewrite, for every $n \in \left\{1,\ldots,N_{\text{train}}\right\}$,
$
\xb_L^{(n)} = \Phi(\btheta, \xb_0^{(n)}),
$
with $\Phi$ a suitable non-linear mapping directly deduced from Eq.~\eqref{eq:bnn1}. If the involved r.v.'s (ouput samples) are assumed to be i.i.d., the unnormalized version of the targeted distribution is given by
\begin{equation}
    \label{eq:target}
\pi (\btheta) =  p\left(\btheta \right) \ell \left( \Y | \btheta, \X_0 \right)
\end{equation}
with
\begin{equation}
\label{eq:loss}
\ell \left( \Y | \btheta, \X_0 \right) = \prod_{n=1}^{N_{\text{train}}} p(\yb^{(n)} | \Phi(\btheta, \xb_0^{(n)})).
\end{equation} 
%\begin{align}
%\Lc \left( \Y | \btheta, \X_0 \right) = \prod_{n=1}^N p \left( \yb^{(n)} | \btheta, \xb_0^{(n)} \right),\\
 %= \prod_{n=1}^N p(\yb^{(n)} | \Phi(\btheta, \xb_0^{(n)})).
%\end{align} 
%\begin{align}
%\pi \left(\btheta | \X_0,\Y \right) & =  p\left(\btheta
%\right)\Lc \left( \Y | \btheta, \X_0 \right),\\
%& =  p\left(\btheta
%\right)\prod_{n=1}^N p \left( \yb^{(n)} | \btheta, \xb_0^{(n)} \right),\\
%& = p\left(\btheta \right) \prod_{n=1}^N p(\yb^{(n)} | \Phi(\btheta, \xb_0^{(n)})).
%\end{align} 
Hereabove, $p\left(\btheta\right)$ is the prior density on the network parameters that will be useful to limit overfitting issues. Training a BNN thus amounts to learning the distribution $\pi(\btheta)$ through sampling and/or variational approximation strategies.  

We now discuss three illustrations of Eq. \eqref{eq:loss} corresponding to standard machine learning scenarios.

%First, the simple regression problem is recovered by considering, for every $n \in \{1,\ldots,N\}$, $\yb^{(n)} \sim \Nc(\xb_L^{(n)}, \sigma^2 \mathbf{I}_{d_{y}})$, with $\sigma > 0$. Robust regression can also be performed, by using generalized Gaussian pdfs instead of the Gaussian one. 
%%
%Second, binary classification consists of setting $d_y = 1$ and using a Bernoulli model $y^{(n)} \sim \Bc(x_L^{(n)})$, assuming that $x_L^{(n)} \in [0,1]$ (this can be easily imposed by an appropriate choice for the output activation function $\Rc_L$). This yields the set cross-entropy as a loss function.
%%
%Third, in classification problems with $C \geq 2$ classes, we have $d_y = C$ and $\yb^{(n)} \in \{0,1\}^C$, which means that the $n$-th ouput is in class $c \in \{1,\ldots,C\}$ if and only if $y_{c}^{(n)} = 1$, while $y_{l}^{(n)} = 0$ for $l \neq c$. The multinomial model $\y^{(n)} \sim \Mc([\x_L^{(n)}]_1,\ldots,[\x_L^{(n)}]_C)$ then leads to the generalized cross-entropy loss.

%
%
%
\subsubsection{Regression problem:} A simple regression problem is recovered by considering, for every $n \in \{1,\ldots,N_{\text{train}}\}$, $\yb^{(n)} \sim \Nc(\xb_L^{(n)}, \sigma^2)$, with $\sigma > 0$, which yields 
\begin{equation}
\label{eq:mse}
\ell \left( \Y | \btheta, \X_0 \right) \propto \exp\left(- \frac{1}{2 \sigma^2}\sum_{n=1}^{N_{\text{train}}} \|\Phi(\btheta, \xb_0^{(n)}) - \yb^{(n)}\|^2\right). 
\end{equation}
From the maximum a posteriori viewpoint, 
this amounts to using a mean square error (MSE) loss in the standard regression context.
A more robust regression can also be performed, by using generalized Gaussian pdfs instead of the Gaussian one. 
\subsubsection{Binary classification problem:} \label{par:binary} Binary classification consists of setting $d_y = 1$ and using a Bernoulli model $y^{(n)} \sim \Bc(x_L^{(n)})$, assuming that $x_L^{(n)} \in [0,1]$. This
range condition can be easily met by a suitable choice for the output activation function $\Rc_L$, e.g. sigmoid. Hence, 
\begin{align}
\ell \left( \Y | \btheta, \X_0 \right) &= \exp\Biggr(\sum_{n=1}^{N_{\text{train}}} \left(y^{(n)} \log\left(\Phi\left(\btheta, \xb_0^{(n)}\right)\right)+ \left(1-y^{(n)}\right)\log\left(1 - \Phi\left(\btheta, \xb_0^{(n)}\right)\right)\right)\Biggr). \label{eq:crossent} 
\end{align}
\subsubsection{General multi-class classification problem:} In classification problems with $C > 2$ classes, we have $d_y = C$ and $\yb^{(n)} \in \{0,1\}^C$, which means that the $n$-th ouput is in class $c \in \{1,\ldots,C\}$ if and only if $y_{c}^{(n)} = 1$, while $y_{\ell}^{(n)} = 0$ for every $\ell \neq c$ (one-hot encoding). The multinomial model $\y^{(n)} \sim \Mc([\x_L^{(n)}]_1,\ldots,[\x_L^{(n)}]_C)$ leads to the generalized cross-entropy training loss:
\begin{equation}
\ell \left( \Y | \btheta, \X_0 \right) = 
\exp\left(\sum_{n=1}^{N_{\text{train}}} \sum_{c=1}^C (y_c^{(n)} \log\left([\Phi(\btheta, \xb_0^{(n)})]_c\right) \right) \label{eq:crossentg}.
\end{equation}
In this case, we must choose $\Rc_L$ so as to satisfy the unit simplex constraints: $\x_L^{(n)} \in [0,1]^C$ and $\sum_{c=1}^C [\x_L^{(n)}]_c = 1$. For example the soft-max activation function can be used.

\subsection{Proposed AIS method}

%The posterior distribution of the unknown parameters in BNNs is intricate due to highly non-linear relationships between the unknown parameters and given data, and also possibly to sophisticated priors. Assuming differentiability of the activation functions and of the prior terms, it is possible to compute the gradient of $- \log \pi (\btheta)$, by making use of the classical back-propagation approach. This feature has been exploited in several Monte-Carlo samplers for BNNs, since it paves the way to Langevin-based acceleration strategies \cite{Wang2018bnn,Nado2018}. In the context of AIS, Langevin scheme has been recently used for proposal adaptation, giving rise to the SL-PMC approach, with promising performance on small to medium scale problems \cite{elvira2019langevin}. However, in the aforementioned work, the scale adaptation was based on the computation of the Hessian matrix of $- \log \pi (\btheta)$, which is barely feasible nor even desirable in the context of BNNs, due to memory limitation issues and mathematical difficulties raised by the models~\cite{BishopHess}. \cred{We thus propose in this paper to resort to another alternative, for covariance matrix adaptation, introduced recently in \cite{el2018robust,ellaham2019recursive}}.

We propose an adaptive importance sampler to deal with the challenges present in the inferential process in BNNs. 
The (unnormalized) posterior distribution of the unknown parameters $\pi(\btheta)$ is intricate due to highly non-linear relationships between the unknown parameters and {the} data, and also possibly to sophisticated priors. 
We will assume differentiability of the activation functions and of the prior terms, so we can compute and exploit the gradient of $- \log \pi (\btheta)$ (via the classical back-propagation approach). Our algorithm belongs to the family of population Monte Carlo (PMC) methods, enhanced by a gradient step in the location parameters update, and a robust and efficient covariance adaptation. Moreover, we introduce a light version of our algorithm which, unlike standard PMC methods, follows a mini-batch strategy, leading to a particularly efficient algorithm, from the viewpoint of both computation and storage.

\begin{algorithm}[!t]
	\centering
	{
	\caption{PMCnet for BNN learning.}
	%\victor{Change to Algorithm instead of Table?}
	\begin{tabular}{p{0.95\columnwidth}}
   % \hline
		\footnotesize
		\begin{enumerate}
			\item {\bf [Initialization]}: 
			\begin{itemize}
			\item[] Load training set $\X_0 \in \Real^{d_{x} \times N_{\text{train}}}$ with output values $\Y \in  \Real^{d_{y} \times N_{\text{train}}}$.
			%\item[] Define the training loss function~\eqref{eq:target}.
			    \item[] Set $\sigma>0$, $(M,K,T) \in (\mathbb{N}^*)^3$, $(\eta_t,\beta_t)_{1 \leq t \leq T}$.
			    \item[] For $m\in \{1,\ldots,M\}$, select the initial adaptive parameters ${\bm \mu}_m^{(1)} \in \mathbb{R}^{d_\theta}$ and $\bSigma_m^{(1)} = \sigma^2 \mathbf{I}_{d_\theta}$.
			\end{itemize}
	 %\cblue{\Real^{d_x \times d_x} }$.% = \sigma^2 \mathbf{I}$.
			\vspace*{6pt}
			\item {\bf[For $\bm t \bm= \bm 1$ to  $\bm T$]}: 
			\begin{enumerate}
				\item Draw $K$ samples from each proposal pdf,
					\begin{equation} 
						\btheta_{m,k}^{(t)} \sim {q_m^{(t)}(\btheta)\equiv \mathcal{N}(\btheta;\bmu_m^{(t)},\bSigma_m^{(t)})},						\label{eq_drawing_part_pmc}
					\end{equation}
					
					with $m\in \{1,\ldots,M\}$, and $k\in\{1,\ldots,K\}$. %and define the set
			%		$$
			%		\mathcal{X}_t=\left\{\x_{i,k}^{(t)}\right\}_{i=1,k=1}^{N,K}.
			%		$$
					\item Compute the importance weights
					\begin{equation} 
						w_{m,k}^{(t)}=\frac{\pi(\thetab_{m,k}^{(t)})}{\frac{1}{M}\sum_{i=1}^M  q_i^{(t)}(\thetab_{m,k}^{(t)})},
					\label{is_part_weights}
					\end{equation}
					with $\pi$ defined in~\eqref{eq:target}.
					%where $\Phi_{i,k}^{(t)}$ is a suitable function (see Section \ref{section_ppmc}).		
					\item Resample $M$ location parameters $\{\widetilde{\bmu}_m^{(t+1)}\}_{m=1}^M$ from the set of $MK$ weighted samples of iteration $t$ {using the local resampling strategy (see \cite{elvira2017improving})}.
				\item Adapt the proposal parameters $\{{\bm \mu}_m^{(t+1)},\bSigma_m^{(t+1)}\}_{m=1}^M$ according to \eqref{eq:AG1} and \eqref{eq:AG2mb}, respectively.
\end{enumerate}
		\item {\bf [Output, $\bm t \bm =\bm T$]}: 
		\begin{itemize}
		    \item[] 	Return the pairs $\{\thetab_{m,k}^{(t)}, {w}_{m,k}^{(t)}\}$, 
				for $m\in\{1,\ldots,M\}$, $k\in\{1,\ldots,K\}$, and $t\in \{1,\ldots,T\}$.
		\end{itemize}
		\end{enumerate} \\
		%\hline 
\end{tabular}\label{PMC_frameworknew}
}
\end{algorithm}

%\jc{It would be very difficult for a non expert in IS to implement the algorithm. Try to be more pedagogical. Especially step~2(c) is unclear.}
%\victor{I agree. I have extended some explanations, but there is room to be more explicit. Not sure if we will have space issues.}

\subsubsection{Description of the algorithm}
Algorithm~\ref{PMC_frameworknew} shows the proposed PMC algorithm for inference in BNNs {that we call PMCnet \cblue{(\textbf{PMC} for Bayesian neural \textbf{net}works)}.} {Without loss of generality and to ease the description, we initialize the algorithm with $M$ Gaussian proposal distributions, $q_{m}^{(1)}(\btheta) \equiv \mathcal{N}\left(\btheta; \bm \mu_{m}^{(1)},\bSigma_{m}^{(1)} \right)$, $m\in \{1,\ldots, M\}$,} with location (i.e., mean) parameters $\{ {\bm \mu}_m^{(1)} \}_{m=1}^M$ and scale (i.e., covariance) parameters $\{ \bSigma_m^{(1)}  \}_{m=1}^M $, that will be adapted iteratively. The algorithm consists of $T$ iterations decomposed into four steps. The sampling step is performed in step~2(a), where $K$ samples are simulated from each proposal pdf. In the weighting step~2(b), each of the $MK$ samples receives an IS weight using the whole mixture of proposals in the denominator (see \cite{elvira2019generalized} for a proof of the associated variance reduction and increased exploratory capabilities). The resampling step is performed in step~2(c), following the {local resampling} of \cite{elvira2017improving}, simulating the set of auxiliary location parameters $\{ \widetilde  {\bm \mu}_m^{(t+1)} \}_{m=1}^M$. {Under this local resampling, the $m$-th parameter $\widetilde{\bm \mu}_m^{(t+1)}$ is resampled from the set of $K$ samples generated by the proposal located at ${\bm \mu}_m^{(t)}$, i.e., from the set 
$\{\btheta_{m,1}^{(t)},\ldots,\btheta_{m,K}^{(t)}\}$
with associated probabilities
$\overline{w}_{m,k}^{(t)}=\frac{w_{m,k}^{(t)}}{\sum_{\ell=1}^{K} w_{m,\ell}^{(t)}},\;k\in \{1,\ldots,K\}$.} This approach guarantees that exactly one sample per proposal survives from $t$ to $t+1$, preserving both diversity and local exploration. In practice, the local resampling then consists in simulating from the categorical distribution $\widetilde{\bm \mu}_m^{(t+1)} \sim \text{Cat}(\{\btheta_{m,k}^{(t)}\}_{k=1}^K ; \{{\overline{w}}_{m,k}^{(t)}\}_{k=1}^K)$ for every $m\in \{1,\ldots,M\}$. \cblue{In step~2(d), a scaled Langevin-based update is performed to update the mean of the proposal density at next iteration as
%\jc{What is the theoretical justification of the use of Langevin equation in this context?}
\begin{equation}
{\bm {\mu}}_{m}^{(t+1)} = \widetilde {\bm \mu}_{m}^{(t+1)} + {\gamma_m^{(t+1)}} \bSigma_{m}^{(t+1)} \nabla \log \pi( \widetilde {\bm \mu}_{m}^{(t+1)} ). \label{eq:AG1}
\end{equation} 
This approach allows to exploit the geometric information
about the target, here through the gradient of $\log \pi$, yielding a more accurate exploration of the target
({see} the theoretical justification of Langevin equation in MC approaches in \cite{Elvira2021,marnissi2018}).}
Hereabove, $\bSigma_{m}^{(t+1)}$ is a symmetric definite positive matrix of $\mathbb{R}^{d_\theta \times d_\theta}$, that will also be used as covariance matrix of the proposal density for next iteration. Moreover, $\gamma_m^{(t+1)} \in (0,1]$ is a stepsize adjusted through a simple backtracking procedure. {To be specific, for each iteration $t$ and each sample $m$, we initialize $\gamma_{m}^{(t+1)}$ to one. We calculate the candidate ${\bm {\mu}}_{m}^{(t+1)}$ using \eqref{eq:AG1} and the related loss value that we compare with the loss of $\widetilde {\bm \mu}_{m}^{(t+1)}$. If the loss decreases, the stepsize is accepted, otherwise we start the backtracking process and try stepsize $\gamma_{m}^{(t+1)}$ reduced by a factor 1/2, until a loss decrease is observed or a maximum number of trials is achieved (typically, 20).} In the SL-PMC scheme \cite{Elvira2021}, the covariance matrix of the proposal density was adapted using $\bSigma_{m}^{(t+1)} = (- \nabla^2 \pi(\widetilde {\bm \mu}_{m}^{(t+1)}))^{-1}$, assuming the inversion is well-defined. We propose instead to {set} $\bSigma_{m}^{(t+1)} $ as a cheaper local approximation of the target curvature, using the following robust covariance estimates, for every $m \in \{1,\ldots,M\}$:
    \begin{equation}
        \label{eq:AG2mb}
       \bSigma_{m}^{(t+1)}  = (1-\beta_t)\bSigma_m^{(t)}+\beta_t(1-\eta_t)\widehat\bSigma_m^{(t)}+\beta_t\eta_t\widetilde\bSigma_m^{(t)}.
    \end{equation}
    Hereabove, $\widehat\bSigma_m^{(t)}$ and $\widetilde\bSigma_m^{(t)}$  are two estimators for the covariance of the target. In practice, we set $\widehat\bSigma_m^{(t)}$ as the empirical covariance computed from the $K$ samples and $K$ associated weights at iteration $t$, and $\widetilde\bSigma_m^{(t)}$ as a biased sample covariance, computed empirically from the $K$ samples from iteration $t$ and a modified version of the  $K$ associated weights, where the $\sqrt{K}$ largest weights values are cropped. This cropping procedure {generally reduces the variance of the importance sampling estimators \cite{ionides2008,koblents2013robust}  (see \cite{martino2018comparison} for a deeper review of these techniques and \cite{vehtari2021pareto} for a similar approach)}. Moreover, in  \eqref{eq:AG2mb}, $0< \beta_t\leq 1$ and $(\eta_t)_{1\le t\le T}$ is a decreasing sequence of constants satisfying $\eta_1=1$, and $\eta_T = 0$. {This convex combination of estimators has been shown to provide more stable updates of the covariance matrix thanks to the incremental estimate approach, governed by $\beta_t$, and also due to the introduction of the lower-variance estimator $\widetilde\bSigma_m^{(t)}$, which is controlled by $\eta_t$.}  This covariance adaptation has been proposed in \cite{ellaham2019recursive}, in the context of AMIS, an AIS family of alternative algorithms to PMC.
    
    %The algorithm outputs $MKT$ weighted samples that can be used to approximate the integral~\eqref{eq_integral}, as we explain hereafter.

\subsubsection{Building the posterior distribution}
\label{posteriorsampler}
{
The proposed {PMCnet} allows to build an approximation to the posterior distribution of the unknown parameters $\btheta$ as
\begin{equation}
\widetilde \pi(\btheta) \approx \sum_{j=1}^J \overline{w}_j \delta (\btheta - \btheta_j),
\label{distribution:theta}
\end{equation}
with $J \geq 1$, where $\{\overline{w}_j,\btheta_j\}_{j=1}^J$ is a subset of the $MKT$ weighted samples $\{{w}_{m,k}^{(t)},\thetab_{m,k}^{(t)}\}_{m,k,t}$, derived from the {PMCnet}, for $m\in\{1,\ldots,M\}$, $k\in\{1,\ldots,K\}${,} and $t\in \{1,\ldots,T\}$. The weights $\overline{w}_j$ are normalized in such a way that $\sum_{j=1}^J \overline{w}_j = 1$. {Larger values of $J$ can lead to a better approximation in \eqref{distribution:theta}, but at the price of a high memory burden.  A typical practical choice (adopted in our experiments) is to define $\{\overline{w}_j,\btheta_j\}_{1 \leq j \leq J}$ as the set of the  samples produced at the last iteration $t=T$, so that $J = MK$.} 
%Given the approximation \eqref{distribution:theta}, it is straightforward to approximate the integral~\eqref{eq_integral}, to yield meaningful moments of the posterior distribution. 
The probabilistic characterization of the learned parameters $\btheta$ through \eqref{distribution:theta} allows us to turn the learned BNN into a generative model, which is of clear interest in multiple data science applications \cite{murphy2012machine}.}

{In the context of machine learning, one of the main goals is the probabilistic prediction of the output of a trained network, given a new test input.} Thanks to our approximation \eqref{distribution:theta}, we can perform this task also in a probabilistic manner. Recall that, in the training set, $N_{\text{train}}$ output observations $\left\{\y^{(n)}\right\}_{1\le n\le N_{\text{train}}} \in \Real^{d_{y}}$ are related to the inputs $\{\xb_0^{(n)}\}_{1\le n\le N_{\text{train}}}$ through $ p(\Y | \Phi(\btheta,\X_0))$. In practice, we obtain only an approximated particle-based version of the posterior of the network parameters, given in \eqref{distribution:theta}. It is possible to simulate from this approximation, an approximation to the distribution of the network response (i.e., output) for any given new test data among $\{\xb_0^{(n)}\}_{ N_{\text{train}}+1 \leq n \leq N_{\text{train}} + N_{\text{test}}}$.
In particular, the propagation of the posterior mean and some metric assessment comparing the ground truth $\left\{\y^{(n)}\right\}_{ N_{\text{train}}+1 \leq n \leq N_{\text{train}} + N_{\text{test}}} \in \Real^{d_{y}}$ to the obtained outputs, can be obtained by the procedure described in Algorithm~\ref{tab:PMCnetoutputmetric}. Note that step (b) simply reduces to $\y^{(r,n)} = \Phi(\btheta^{(r)},\x_0^{(n)})$ for the three examples of likelihood presented in Section~\ref{sec:proposedmodel}. 
\begin{algorithm}[]
    \centering
   \begin{tabular}{p{0.95\columnwidth}}
   % \hline
        1. \textbf{[Initialization]} \cblue{Select $J$ pairs $\{\btheta_j,w_j\}_{1 \leq j \leq J}$  among $MKT$ \cblue{weighted} samples $\{\btheta_{m,k}^{(t)},w_{m,k}^{(t)} \}_{m,k,t}$. One can select for instance only the $MK$ samples of the last iteration, $T$, or the whole set of $MKT$ samples (see \cite{bugallo2017adaptive,elvira2021advances} for a discussion on the estimators in IS). Then, compute the normalized weights  
        $\big\{\overline{w}_j = {w}_j/\sum_{i=1}^J {w}_i\big\}_{j=1}^J$
        %$\{\overline{w}_j\}_{1 \leq j \leq J}$ 
        of the selected $J$ pairs.}
        %, with $\overline{w}_j = {w}_j/\sum_{i=1}^J {w}_i$.}
        \\
   2. \textbf{[For $r = 1,\ldots,R$]}
\begin{enumerate}
\item[(a)] {Sample} $\btheta^{(r)}$ from the set $\{\btheta_j \}_{j=1}^J$, with probability $\{\overline{w}_j \}_{j=1}^J$,
\item[(b)] {Evaluate} $\y^{(r,n)} = \mathbb{E}(\y | \Phi(\btheta^{(r)},\x_0^{(n)}))$ for $n \in \{N_{\text{train}}+1,\ldots,N_{\text{train}} + N_{\text{test}}\}$,
\item[(c)] Compute the metric of interest, comparing the outputs $\{\y^{(r,n)}\}_{N_{\text{train}}+1 \leq n \leq N_{\text{train}} + N_{\text{test}}}$ to ground truth values $\{\y^{(n)}\}_{N_{\text{train}}+1 \leq n \leq N_{\text{train}} + N_{\text{test}}}$.
\end{enumerate}\\
% \hline
    \end{tabular}
    \caption{Output computation and metric assessment of PMCnet network on a test set $\{\widetilde{\x}_0^{(n)}\}_{N_{\text{train}}+1 \leq n \leq N_{\text{train}} + N_{\text{test}}}$. }
    \label{tab:PMCnetoutputmetric}
\end{algorithm}

\subsubsection{Practical implementation}

\paragraph{Code availability}
For reproducibility {purposes}, we share a repository {available at \url{https://github.com/yunshihuang/PMCnet}} with our implementation in PyTorch of the proposed algorithm. The advantage of a framework such as PyTorch is that it contains many built-in functions to deal with nonlinear feedforward neuronal structures such as as \eqref{eq:bnn1}. These functions make {an efficient} use of the available GPU resources. Auto-differentiation tools are also used to compute the gradient of our target function, without the need for any explicit tedious calculations. In our code, we define the parameters $\thetab$ (i.e., weights/biases of the network) as Pytorch tensors through the option \texttt{requires\_grad=True}. This allows us to compute the loss and its {gradient} in parallel (using GPU) for a given list of weights/biases values, using \texttt{backward} PyTorch function. For instance, the output and the associated target value for $M$ trial samples in our algorithm can be simply evaluated simultaneously, without {the} need for a loop. Despite these advantages, when the dimension $d_{\theta}$ of the unknown parameters and/or the size of the training set $N_{\text{train}}$ increase, the memory cost of the algorithm might still be high. 

\paragraph{\cblue{PMCnet-light algorithm}} \label{sec:pmcnetlight}
We thus propose a modified version of {PMCnet}, that {avoids} memory overflows without being detrimental to the numerical performance of the method (as we will illustrate in our experimental section). Two changes are done, leading to  {PMCnet-light}.\\
First, to cope with large values of $d_\theta$, we propose to modify the definition of the adapted covariance matrix $\bSigma_m^{(t+1)}$ in {Eq.~\eqref{eq:AG2mb}}, using a diagonal scaling instead. For every iteration $t \in \{1,\ldots,T\}$ and {proposal} $m \in \{1,\ldots,M\}$, we define
\begin{align}
\bDelta_{m}^{(t+1)}  = (1-\beta_t)\bDelta_m^{(t)}+\beta_t(1-\eta_t)\widehat\bDelta_m^{(t)}+\beta_t\eta_t \widetilde\bDelta_m^{(t)}.
\label{eq:AG2mb_diag}   
\end{align}
Hereabove, $\bDelta_{m}^{(t+1)}$, $\widehat\bDelta_m^{(t)}${,} and $\widetilde\bDelta_m^{(t)}$ are diagonal matrices of $\mathbb{R}^{d_\theta \times d_\theta}$. Matrices $\widehat\bDelta_m^{(t)}$ (resp. $\widetilde\bDelta_m^{(t)}$) are built in such a way that its diagonal elements match with those of the previously defined $\widehat\bSigma_m^{(t)}$ (resp. $\widetilde\bSigma_m^{(t)}$). Parameters $(\beta_t,\eta_t)$ play the same role as in the original version of the algorithm. Using diagonal matrices here offers {the advantages of (i) a reduced memory load, and (ii) a computational complexity decrease in the sampling step in \eqref{eq_drawing_part_pmc}}. \\
Second, to tackle large values of $N_{\text{train}}$, we introduce an incremental gradient strategy in our adaptation rule~\eqref{eq:AG1}. The idea is to approximate the full batch gradient involved in~\eqref{eq:AG1} by one loop (i.e., one epoch) of mini-batch gradient steps. To be specific, let us divide our training set $\{\xb^{(n)},\yb^{(n)}\}_{1 \leq n \leq N_{\text{train}}}$ into $B$ batches of equal size $N_{\text{train}}/B$, and denote, for every batch index $b \in \{1,\ldots,B\}$, $\X_b \in \mathbb{R}^{d_x \times \frac{N_{\text{train}}}{B}}$ and $\Y_b \in \mathbb{R}^{d_y \times \frac{N_{\text{train}}}{B}}$, the column-wise concatenation of $\{\xb^{(n)}\}_{N_{\text{train}} (b-1)/B+1 \leq n \leq N_{\text{train}} b/B}$ and $\{\yb^{(n)}\}_{N_{\text{train}} (b-1)/B+1 \leq n \leq N_{\text{train}} b/B}$, respectively. Then, using \eqref{eq:loss}, for every $\thetab \in \mathbb{R}^{d_\theta}$, we have:
$
    \pi(\thetab) = \prod_{b = 1}^{B} \pi_b(\thetab),
$
with
\begin{equation}
    \pi_b(\thetab) = (p(\theta))^{1/B} \ell(\Y_b | \theta,\X_b).
\end{equation}
{For any} iteration index $t \in \{1,\ldots,T\}$ and a sample index $m \in \{1,\ldots, M\}$, we set,
\begin{align}
& \widehat{\bm{\mu}}_{m,1}^{(t+1)}= \widetilde{\bm{\mu}}_{m}^{(t+1)} \notag\\
& \text{For }b=1,\ldots,B \notag\\
& \left \lfloor \quad \widehat{\bm{\mu}}_{m,b+1}^{(t+1)} = \widehat{\bm{\mu}}_{m,b}^{(t+1)} + \frac{\gamma_{m,b}^{(t+1)}}{2} \bDelta_{m}^{(t+1)} \nabla \log \pi_b( \widehat{\bm{\mu}}_{m,b}^{(t+1)} ) \right., \label{eq:AG1_mini}\\
& {\bm {\mu}}_{m}^{(t+1)} = \widehat{\bm{\mu}}_{m,B+1}^{(t+1)}. \notag
\end{align}
Hereabove, $\gamma_{m,b}^{(t+1)}$ is a stepsize that is adjusted through the backtracking process on the mini-batch loss $\pi_b$, similarly to what was done for the full batch version of the algorithm. The diagonal version of the covariance matrix of the proposal $\bDelta_{m}^{(t+1)}$, defined in \eqref{eq:AG2mb_diag}, is kept the same for all batches. The proposed approach can be viewed as running one epoch of an incremental (scaled) gradient strategy. Since {the mean parameters} are {updated $B$ times} per iteration and per sample, {this adaptation allows for a better exploration of the target}. Moreover, this strategy is also less computationally demanding since each mini-batch data can be loaded on the fly, without being fully stored. 

\cblue{In a nutshell, the resulting {PMCnet-light} amounts to applying Algorithm \ref{PMC_frameworknew}, where 
only Step 2(d) is modified.
%the only change lies in Step 2(d). 
The proposal parameters $\{{\bm \mu}_m^{(t+1)},\bDelta_m^{(t+1)}\}_{m=1}^M$ adaptation is now performed according to \eqref{eq:AG1_mini} and \eqref{eq:AG2mb_diag}, respectively.
}

\section{Numerical experiments}
\label{sec:experiments}
We now illustrate the good performance of {PMCnet} and its low complexity version {PMCnet-light}, on a bunch of classification and regression problems involving either shallow NNs or DNNs.

\subsection{Numerical settings}

\subsubsection{Architecture design}
For binary classification problems (i.e., $C=2$), we consider the likelihood \eqref{eq:crossent}  associated with the cross-entropy loss function while, for multi-class classification (i.e., $C>2$), we consider the one associated with the generalized cross-entropy loss in  \eqref{eq:crossentg}. For regression tasks, the Gaussian likelihood corresponding to the MSE loss \eqref{eq:mse} is used. Concerning the network structure, FCNNs make use of either hyperbolic tangent ($tanh$) or rectified linear unit (ReLU) activation functions $\mathcal{R}_{\ell}$ in \eqref{eq:bnn1}, for $\ell \in \{1,\ldots,L-1\}$, as precised hereafter for each example. In binary classification problems, $\mathcal{R}_{L}$ is the sigmoid function whereas, for multi-class classification, we set $\mathcal{R}_{L}$ to the softmax function. For regression, $\mathcal{R}_{L}$ reduces to identity. {In all experiments, for simplicity, we choose as a prior distribution of each entry of the unknown parameter $\btheta$ an i.i.d. zero-mean Gaussian distribution with constant variance manually finetuned for each network (see more details below)}.

%\ccyan{For $t=1,\ldots, T$, the parameters of the robust covariance adaptation in our RSL-PMC are set as $\beta_{t} \equiv 0.5$, $\eta_{t}\equiv \frac{1}{t}$, both $\gamma_{m}^{(t+1)}$ and $\gamma_{m,b}^{(t+1)}$ are initialized as 2 and the multiplied factor is 0.5 during the backtracking process.} In all the tests, we split the dataset into three different parts: 1) training set used for running the inference algorithms, 2) validation set used for choosing the optimal regularization weight related to the parameter prior and the optimal iterations $T$, and 3) test set used for evaluating quantitatively the performance of the trained model. If unspecified, we only list the sample size of the whole dataset and the proportion of these three sets is set as 6:2:2. Otherwise we would provide the sample size of each set respectively.

\subsubsection{Comparisons to other methods} \label{sec:bench}
Various methods are considered as comparison benchmarks. First, we provide the results obtained by a standard (i.e., non Bayesian) training procedure relying on ADAM optimizer. We use ADAM to either compute the minimizer of the neg-log-likelihood $-\log \mathcal{L}(\textbf{Y}|\theta, \textbf{X}_{0})$, leading to the maximum likelihood solution denoted ADAM-MLE, or to compute the maximum a posteriori solution ADAM-MAP defined as the minimizer of $-\log (p(\theta)\mathcal{L}(\textbf{Y}|\theta, \textbf{X}_{0}))$.  {The estimated parameters obtained by ADAM-MLE are used as the initialization of our {PMCnet} \cblue{(or PMCnet-light)}.} Then, several state-of-the-art Bayesian neural networks approaches are evaluated, namely Bayes by Backprop (BBP) \cite{Blundell2015}, Stochastic Gradient Langevin Dynamics (SGLD) \cite{WellingSGLD}, MCDropout \cite{Gal2016}, and {Sequential Anchored Ensembles (SAE) \cite{Arnaud2022}}. BBP uses unbiased estimates of gradients of the cost to learn a variational posterior distribution over the weights. We always choose the diagonal Gaussian distribution as the variational posterior for each weight/bias. SGLD adds noise to a standard stochastic gradient optimization algorithm, with annealing stepsize, to push the iterates towards samples from the sought posterior distribution. MCDropout drops a unit with certain probability, and thus models the variability with dropout NN models. {SAE trains an ensemble of models based on anchored losses, by using a guided walk
Metropolis-Hastings procedure with Gaussian transitions, with the aim to provide an estimate of the Bayesian posterior. This method achieved the 2nd (resp. 3rd) place in the light (resp. extended) track of the NeurIPS 2021 Approximate Inference in Bayesian Deep Learning competition.} {To finetune the hyperparameters for these competitors, we always pick the optimal values based on their respective performance on the validation set of each example. For most competitors, we rely on the implementations available at \url{https://github.com/JavierAntoran/Bayesian-Neural-Networks}. We thank the authors of SAE for sharing their code.}

\subsubsection{Evaluation metrics}
The performance of each compared methods is quantified using several standard machine learning metrics, which are evaluated on the test set. For classification tasks, we provide accuracy and confusion matrix, defining the predicted class as the one maximizing the model output. For binary classification problems, we are also able to compute precision, recall, specificity, and F1 score, using a threshold of $0.5$ on each network output to determine the classification decision. For methods capable of computing probabilistic estimates, we evaluate the performance in terms of mean and standard deviation {denoted as std}. Precisely, for all the benchmark methods providing probabilistic estimates (namely, SGLD, MCDropout, and {SAE}), we rely on the strategies described in their seminal papers, to compute the mean, standard deviation and confidence intervals, for the metrics computed on the test set. For our method PMCnet (and its variant PMCnet-light), we rely on the procedure proposed in Table~\ref{tab:PMCnetoutputmetric}. For the sake of readibility, only the mean values are provided in the confusion matrices. {As for multi-class classification problems, we compute F1 score averaged over all the classes.} For regression tasks, we provide the mean (std, if available) of the mean squared error (MSE) for each method. To further assess the performance of different methods in the classification examples, we also display ROC plots (i.e., false vs true positive rate, for varying threshold values from $0$ to $1$ defining the predicted class) and compute the associated area-under-curve value (AUC). In multi-class case, a one-versus-all approach is used to define the ROC and AUC per class. For methods that are able to provide probabilistic estimations, we additionally display some ROC envelopes associated to specific credible intervals (CI) in percentage (95\% CI, if not specified otherwise) of the false vs true positive rate.  

\subsubsection{Training specifications}
In all the experiments, we split the dataset into three different parts: (1) training set used for running the inference algorithms, (2) validation set used for choosing the optimal regularization weight related to the parameter prior and the optimal iteration number $T$, and (3) test set used for evaluating quantitatively the performance of the trained models. Unless otherwise specified, the proportion of the train/validation/test is set as 6:2:2, with $N_{\text{train}}$ (resp. $N_{\text{test}}$) denoting the number of examples in the training (resp. validation or test) sets. {For all the methods, the train/validation/test phases are implemented in Pytorch (version 1.7.0) under Python (version 3.6.10) environment, and run on an Nvidia DGX workstation using one Tesla V100 SXM2 GPU (1290 MHz
frequency, 32GB RAM). The code of our method is made available at {\url{https://github.com/yunshihuang/PMCnet}} for reproducibility purposes.}

\subsubsection{{PMCnet} tuning}
%\emilie{VE, please check carefully this subsection, thanks !}
Regarding the settings of {PMCnet} (or {PMCnet-light}), {for $t=1,\ldots, T$, the parameters of the robust covariance adaptation are set to $\beta_{t} \equiv 0.5$, $\eta_{t}\equiv \frac{1}{t}$. {Both} $\gamma_{m}^{(t+1)}$ and $\gamma_{m,b}^{(t+1)}$ are initialized as $1$ and multiplied by 0.5, sequentially, during the backtracking trial procedure.} {Unless otherwise stated, we set $(M,K)=(50,100)$.} Since the regularization weight and the iteration number $T$ are essential parameters that can affect the performance of our method, we adopt the following strategy to finetune these hyperparameters. We first run our method on the training set for a relatively large number of iterations (typically, $T = 70$), and run Algorithm \ref{tab:PMCnetoutputmetric} on the validation set, using the $J = MK$ samples of the last iteration $T$ and $R = 100$. \cblue{It would be possible to use all samples, i.e.,  $J = TMK$ samples, although we have observed that in practice it does not make a significant difference.} %\ccyan{By choosing a relatively big $T$, it ensures that those learnt location parameters at the last iteration converges eventually to the true parameters.} 
This allows to calculate the averaged accuracy (or averaged MSE for regression task) on the   validation set. The variance parameter involved in the prior distribution is set by {golden} search so as to {maximize the averaged accuracy for classification task (resp. minimize the averaged MSE for regression task)}. A similar strategy is adopted to set up the optimal iteration number $T$. Namely, once the regularization weight has been set, we try different values for $T$, and define it as the smallest value (thus, smallest complexity) allowing to reach stable performance (averaged accuracy or MSE) on the validation set. 

{Once hyperparameters are set, we run the procedure of section \ref{posteriorsampler} on the test set, again using $R = 100$, which provides us the mean, variance, and CIs for metrics of interest (using Algorithm \ref{tab:PMCnetoutputmetric}) as well as distribution plots for the predicted output values on test set examples.}

\subsection{Experimental results}

\subsubsection{Performance assessment on a control scenario}
\label{secshallow}
We first evaluate our approach {PMCnet} and its variant PMCnet-light on a control dataset mimicking a binary classification problem. No validation set is used in these simple examples, and we always set $(M,K,T)=(50,100,20)$.

%on two small datasets of binary classification, in order to assess the validity of the proposed procedure. We consider a control dataset and the dataset \emph{ionosphere} from LIBSVM library. We consider three different train/test splits to better illustrate the probabilistic inference feature of our method. No validation set is used in these simple examples, and we always set $(M,K,T)=(50,100,20)$.

The considered shallow FCNN is described in Table~\ref{table:network_syn}. The ground truth labels for the control dataset are generated by setting a given (shallow) FCNN architecture for which the groundtruth vector $\bar{\thetab}$ is known. Specifically, we design a synthetic FCNN with $L = 2$ layers, with $tanh$ as the activation function for the hidden layer, and random weight and biases entries independently drawn from $ \mathcal{N}(0,2^{2})$ for every layer. We then feed the network with random $N_{\text{train}}$ (resp. $N_{\text{test}}$) inputs entries $\{\textbf{x}_{0}^{(n)}\}_{1 \leq n \leq N_{\text{train}}}$ (resp. $\{\widetilde{\textbf{x}}_{0}^{(n)}\}_{N_{\text{train}}+1 \leq n \leq N_{\text{train}}+ N_{\text{test}}}$)  independently drawn from $ \mathcal{N}(0,1)$. This yields  $N_{\text{train}}$ (resp. $N_{\text{test}}$) associated ground truth output values $\{\textbf{y}^{(n)}\}_{1 \leq n \leq N_{\text{train}}}$ (resp. $\{\widetilde{\textbf{y}}^{(n)}\}_{N_{\text{train}}+1 \leq n \leq N_{\text{train}}+ N_{\text{test}}}$ ) used for loss evaluation at training (resp. metrics evaluation at testing) phases. 

\begin{figure}[H]
\hspace{-3cm}
\begin{tabular}{@{}c@{}c@{}c@{}}
\includegraphics[width = 6.5cm,height = 5cm]{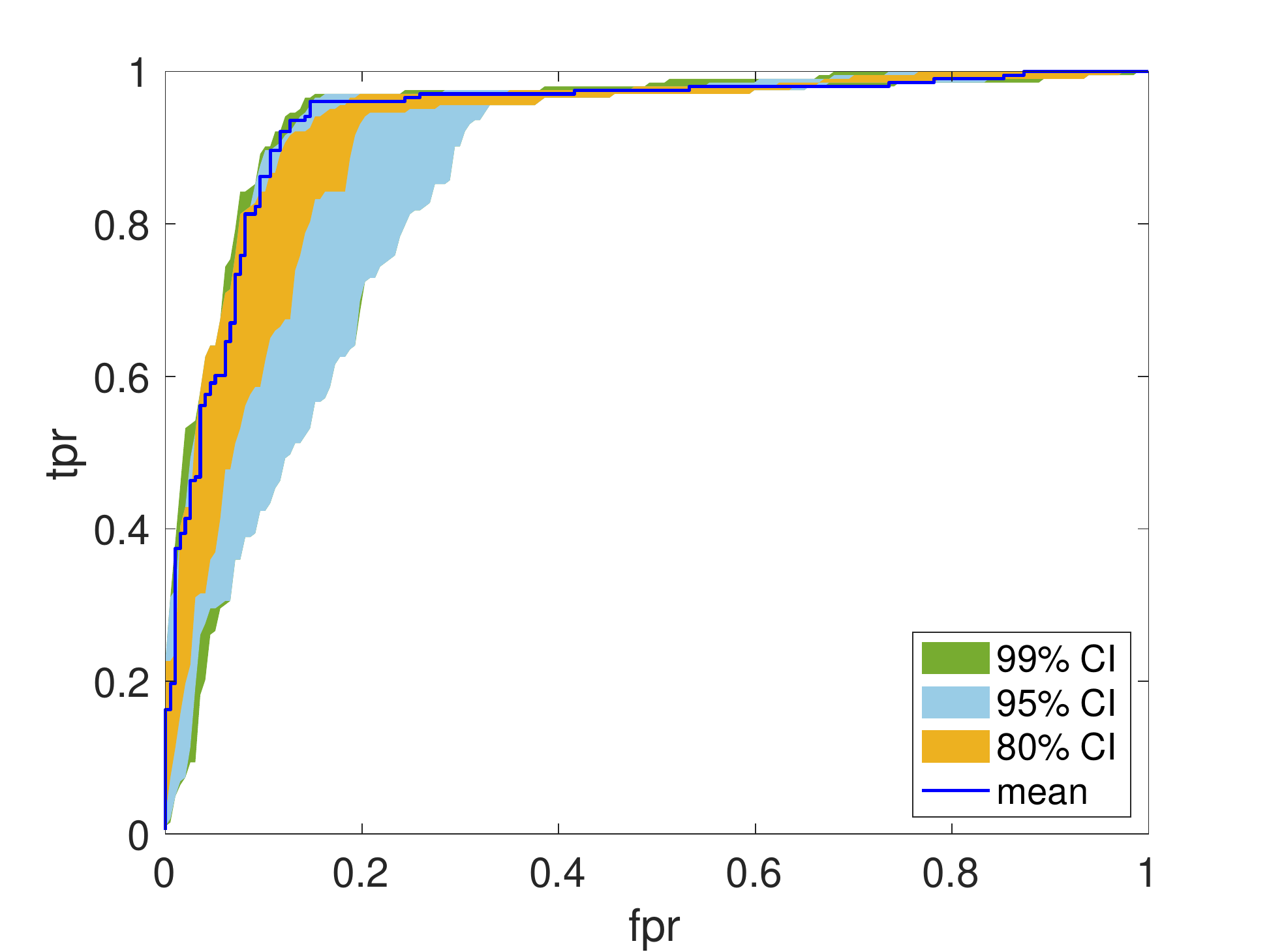}& \hspace{-0.5cm}
\includegraphics[width = 6.5cm,height = 5cm]{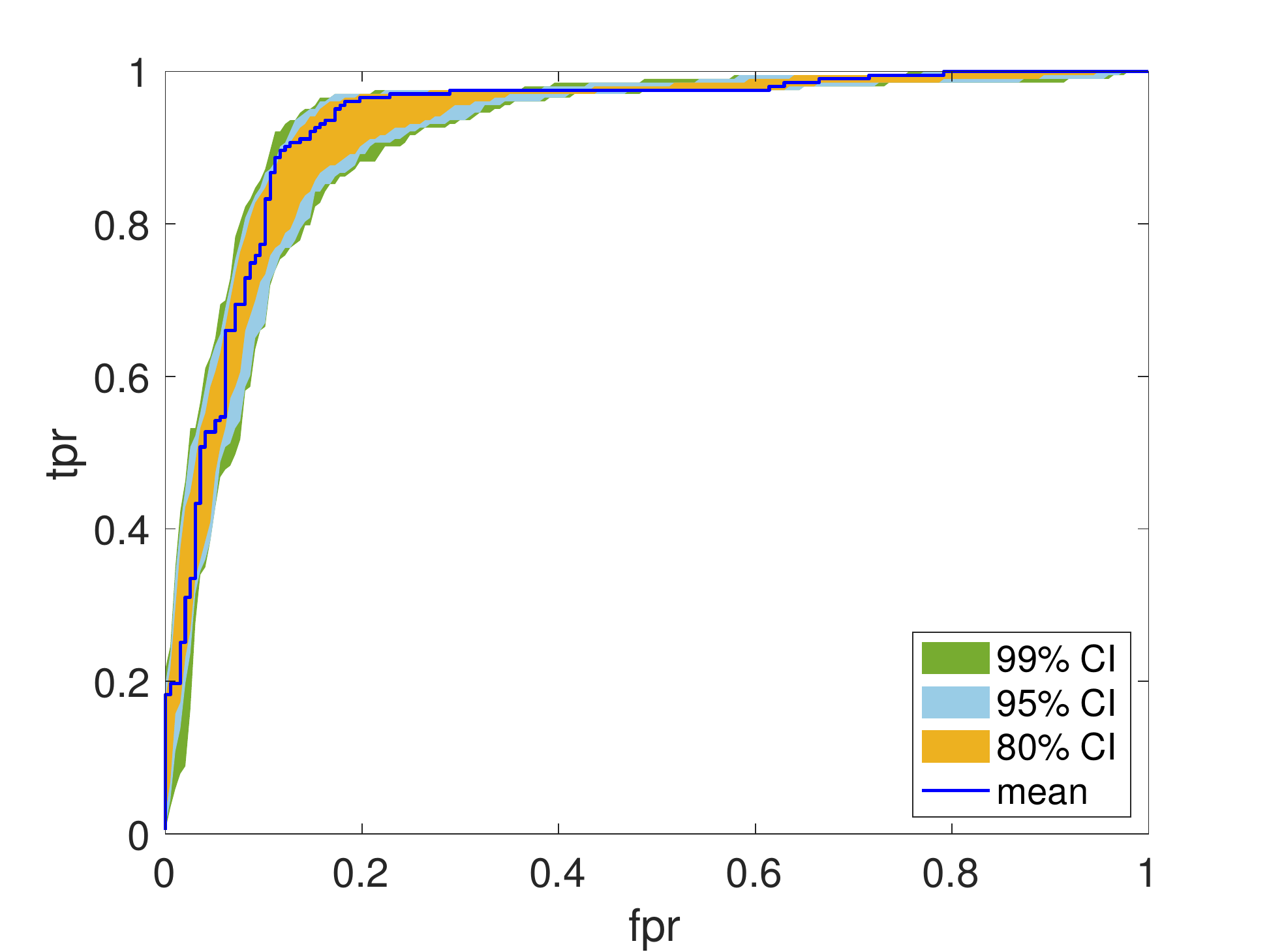}& \hspace{-0.5cm}
\includegraphics[width = 6.5cm,height = 5cm]{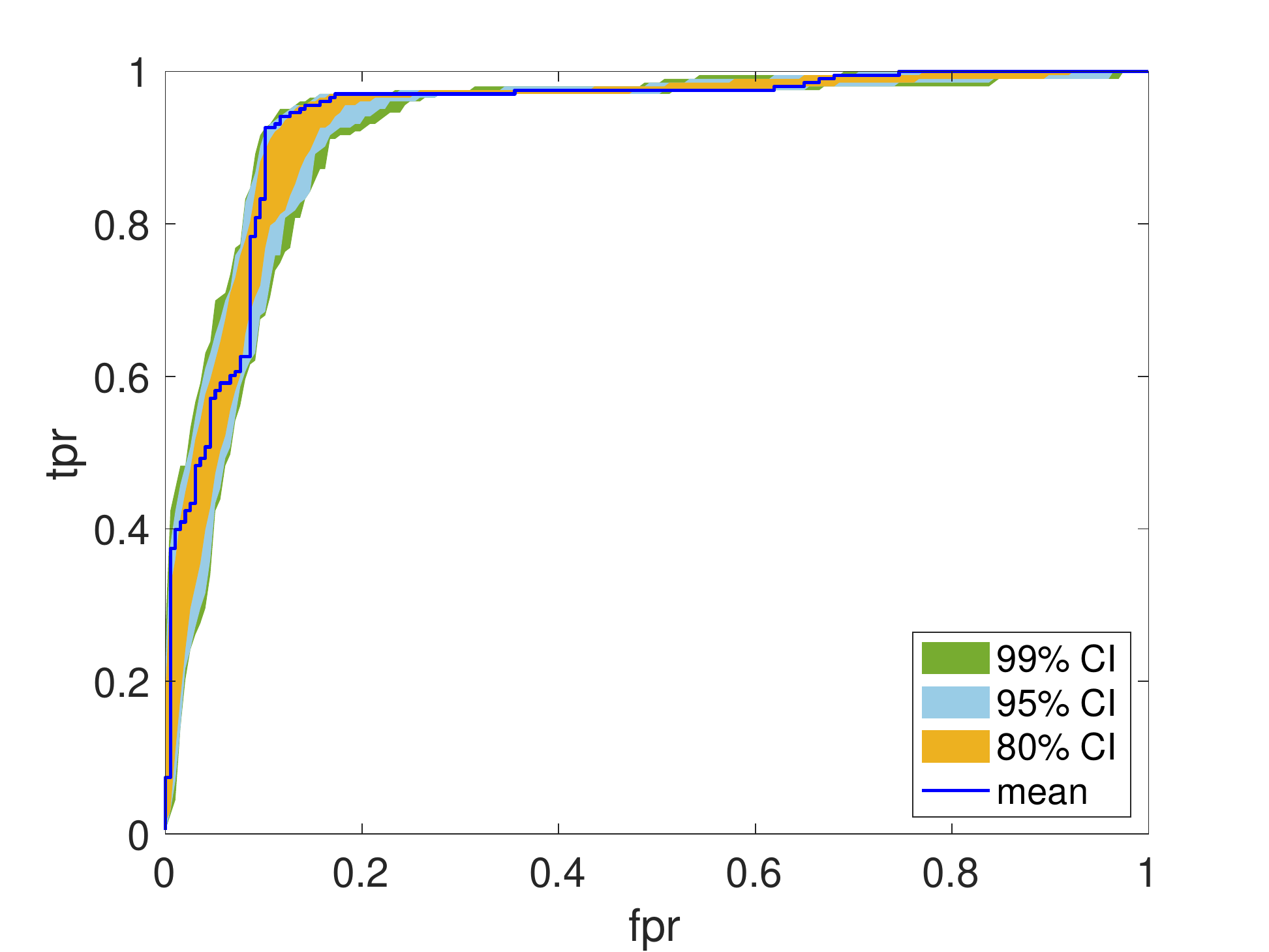}  \\
(a) & (b) & (c)\\
\end{tabular}
\caption{\footnotesize ROC curves of {PMCnet} on test set, on control dataset using (a) $N_{\text{train}} = 50$, (b) $N_{\text{train}} = 400$ and (c) $N_{\text{train}} = 1600$ with 80\% CI, 95\% CI and 99\% CI respectively.}
\label{ROC:syn_CI_all}
\end{figure}

We first investigate the influence of the train/test split on PMCnet results. We set $N_{\text{test}} = 400$, and run experiments with either $N_{\text{train}}= 50, 400$, or $1600$. 
%For \emph{ionosphere}, we set $N_{\text{test}} = 85$, and perform training using either $N_{\text{train}}= 25, 85$ or $255$. 
We report the resulting classification metrics on the test set in Table~\ref{table:result_syn}, for these different values of $N_{\text{train}}$. The performance is good, showing that the method is learning the model in a suitable way, and as expected, the larger training set, the better the classification metrics. Interestingly, the variability of the output metrics obtained by our method is growing as the train/test split is less favorable. Similar conclusions are reached by inspecting the ROC curves and their CI envelopes in Fig.~\ref{ROC:syn_CI_all}. The envelopes are wider (i.e., exhibit more variability) for smaller train sets, showing the interest and relevance of the provided posterior estimation. 

\begin{table}[H]
\footnotesize
\centering
\begin{tabular}{|c|c|c|c|c|c|}
\hline
Number of & Number of & Input  & Output& Number of & Number of \\ 
layers $L$ & classes $C$& size $S_{0}$ & size $d_{y}$  & hidden layers $S_{1}$ & parameters $d_{\theta}$ \\
\hline
2 & 2 & 3& 1 & 3& 16\\
\hline
\end{tabular}
\vspace{0.5cm}
\caption{\small Settings of the FCNN architectures for the control dataset.}
\label{table:network_syn}
\end{table}

% \begin{table}[H]
% \footnotesize
% \centering
% \begin{tabular}{|c||c|c|c|c|c|c|}
% \hline
% Dataset & Number of & Number of & Input  & Output& Number of & Number of \\ 
% name & layers $L$ & classes $C$& size $S_{0}$ & size $d_{y}$  & hidden layers $S_{1}$ & parameters $d_{\theta}$ \\
% \hline
% Control & 2 & 2 & 3& 1 & 3& 16\\
% \hline
% Ionosphere & 2 & 2 & 33& 1 & 3& 106\\
% \hline
% \end{tabular}
% \vspace{0.5cm}
% \caption{\small Settings of the FCNN architectures for the control dataset and dataset \emph{Ionosphere}.}
% \label{table:network_syn}
% \end{table}

\begin{table}[H]
\hspace{-2cm}
\footnotesize
\begin{tabular}{|c|c|c|c|c|c|c|c|}
\hline
$N_{\text{train}}$ & \textbf{AUC} & \textbf{Precision} & \textbf{Recall} & \textbf{Specificity} & \textbf{Accuracy} & \textbf{F1 score} & \textbf{Confusion matrix}\\ 
\hline
50 &0.9183 (0.0191) &0.9347 (0.0216) & 0.7087 (0.1294) & \textbf{0.9493} (0.0264)&0.8308 (0.0525) &0.7984 (0.0767)  & $\begin{bmatrix}
140 &57\\
10& 193
\end{bmatrix}$   \\ 
\hline
400 &0.9234 (0.0070) &0.8946 (0.0364) & \textbf{0.8412} (0.0233) &0.9017 (0.0394) &0.8719 (0.0161) &0.8663 (0.0148) & $\begin{bmatrix}
166 &31\\
20& 183
\end{bmatrix}$   \\ 
\hline
1600 &\textbf{0.9304} (0.0051)&\textbf{0.9410} (0.0196) &0.8396 (0.0202)  & 0.9483 (0.0193) & \textbf{0.8948} (0.0089) &  \textbf{0.8871} (0.0098) &$\begin{bmatrix}
165 & 32\\
10 &193
\end{bmatrix} $ \\
\hline
\end{tabular}
\caption{\small Results of {PMCnet} on test set, for binary classification on control dataset using different values for $N_{\text{train}}$.}
\label{table:result_syn}
\end{table}

\cblue{
We now illustrate the performance of PMCnet when compared to four variants of it,  on the same control dataset, where we now set $N_{\text{train}}=1600$. Namely, in the spirit of an ablation study, we compare PMCnet with
\begin{itemize}
    \item a gradient-free implementation, PMCnet-gradfree, where we discard the Langevin-based mean adaptation \eqref{eq:AG1}, by setting $\gamma_m^{(t)} \equiv 0$ in Algorithm~\ref{PMC_frameworknew};
    \item a covariance-free implementation, PMCnet-covfree, where we discard the covariance adaptation \eqref{eq:AG2mb}, by setting $\bSigma_m^{(t)} \equiv \sigma^2 \mathbf{I}_{d_\theta}$;
    \item the DM-PMC algorithm \cite{elvira2017improving}, where both mean and covariance adaptation are discarded;
    \item the low complexity variant PMCnet-light described in Section~\ref{sec:pmcnetlight}.
\end{itemize}
 }

%\emilie{todo EC: add the results and comment}

\begin{figure}[H]
\hspace{-3cm}
\begin{tabular}{@{}c@{}c@{}c@{}}
\includegraphics[width = 6.5cm,height = 5cm]{ROC_syn_1600_CI_all_new.pdf}& \hspace{-0.5cm}
\includegraphics[width = 6.5cm,height = 5cm]{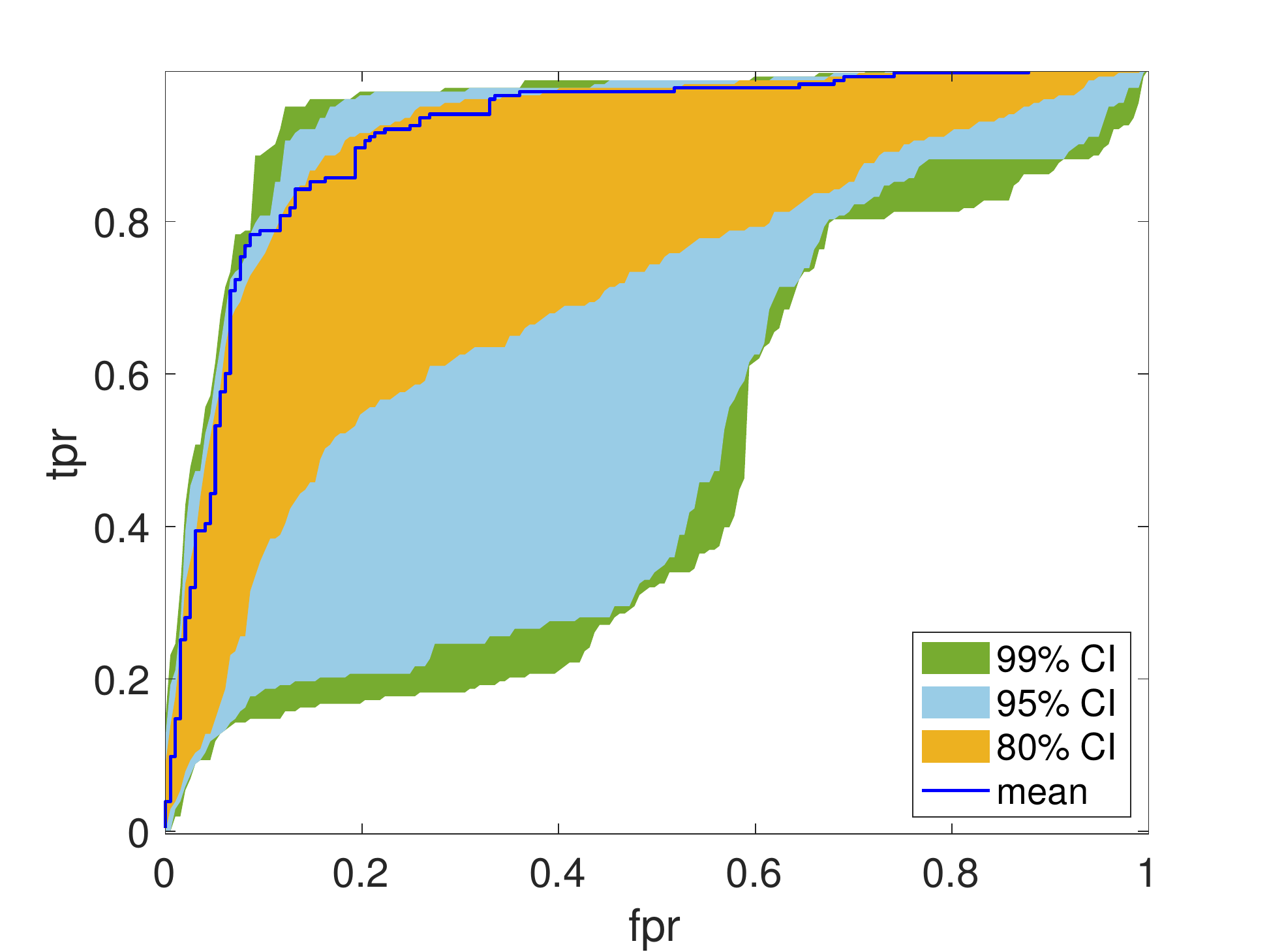}& \hspace{-0.5cm}
\includegraphics[width = 6.5cm,height = 5cm]{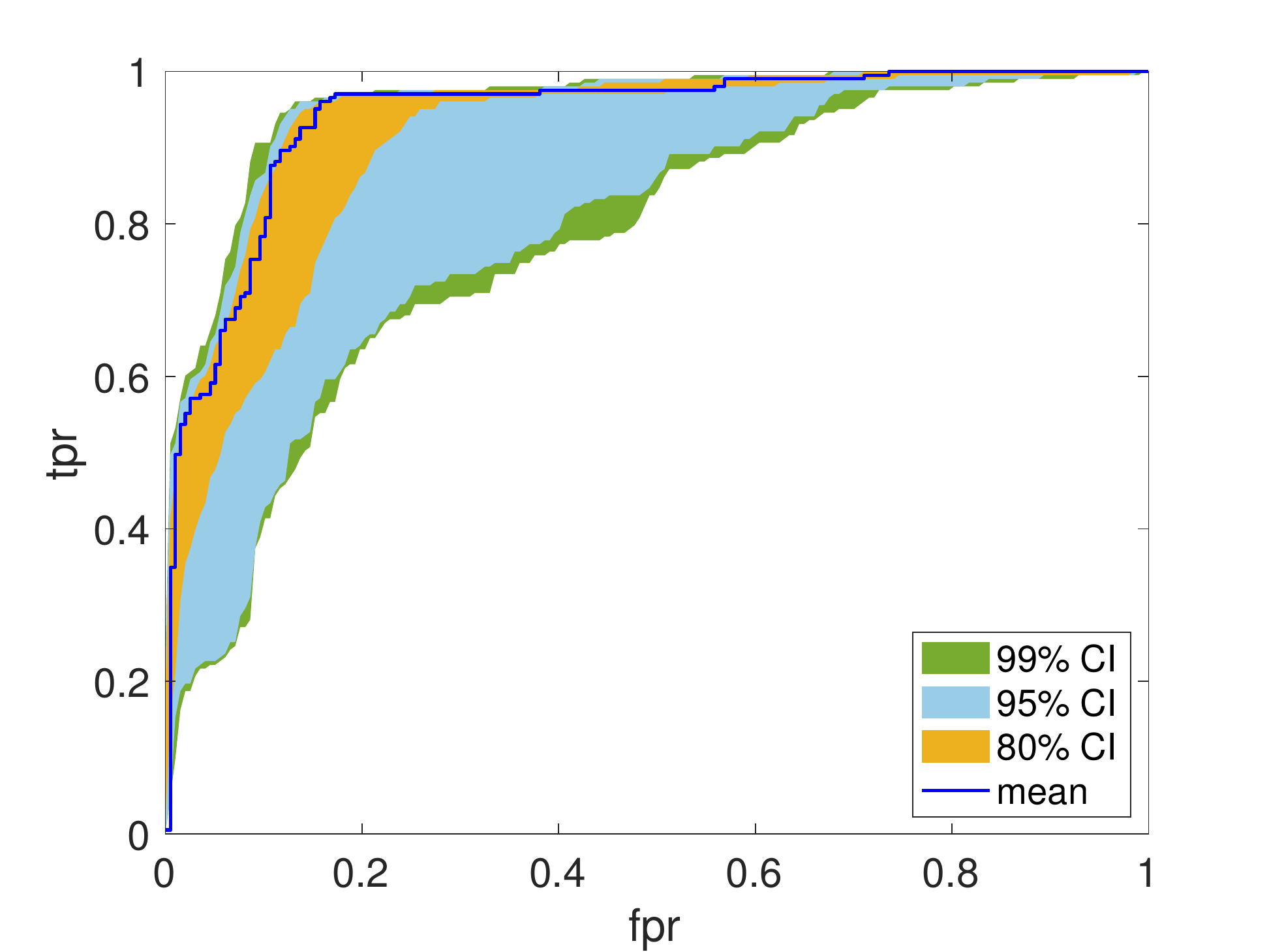}  \\
(a) & (b) & (c)\\
\end{tabular}
\begin{tabular}{@{}c@{}c@{}}
\includegraphics[width = 6.5cm,height = 5cm]{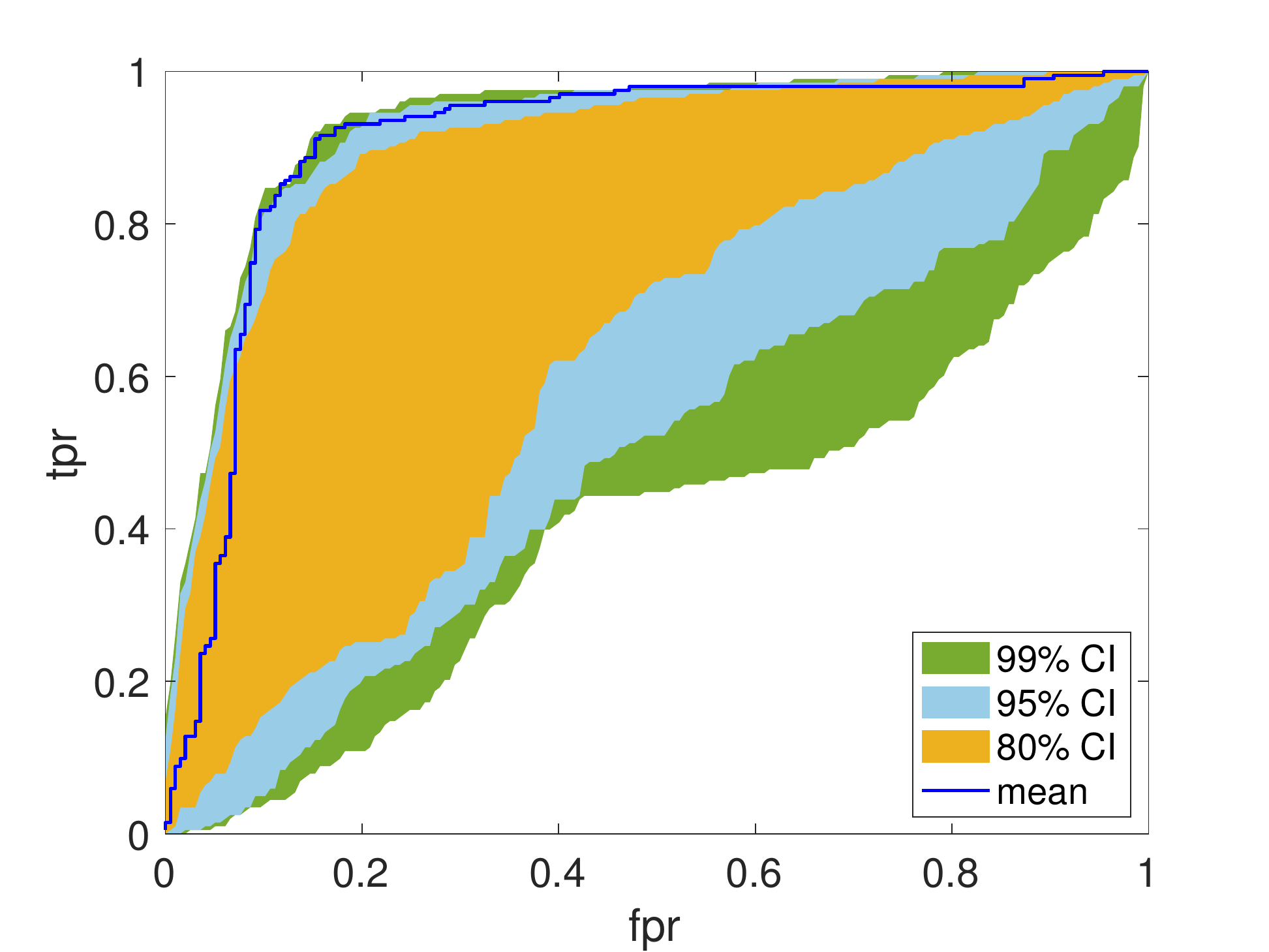}& \hspace{-0.5cm}
\includegraphics[width = 6.5cm,height = 5cm]{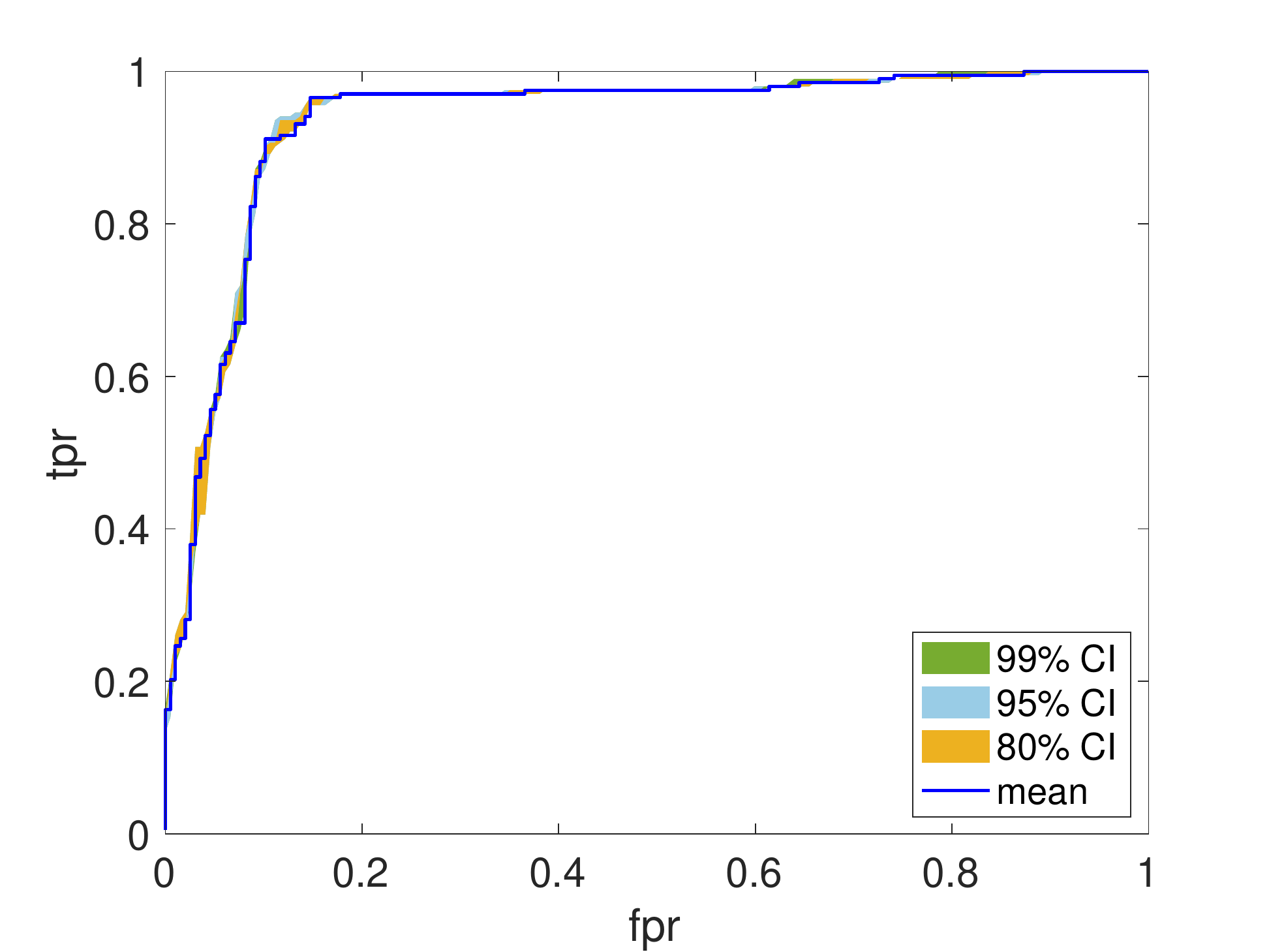}\\
(d) & (e) \\
\end{tabular}
\caption{\footnotesize \cblue{ROC curves (mean and CIs) on test set, using control dataset with $N_{\text{train}} = 1600$, with (a) {PMCnet}, (b) PMCnet-gradfree, (c) PMCnet-covfree, (d) DM-PMC, and (e) {PMCnet-light}.}}
\label{ROC:syn_CI_all_diag}
\end{figure}

\begin{figure}[H]
\begin{tabular}{@{}c@{}c@{}}
\includegraphics[width = 6.5cm,height = 5cm]{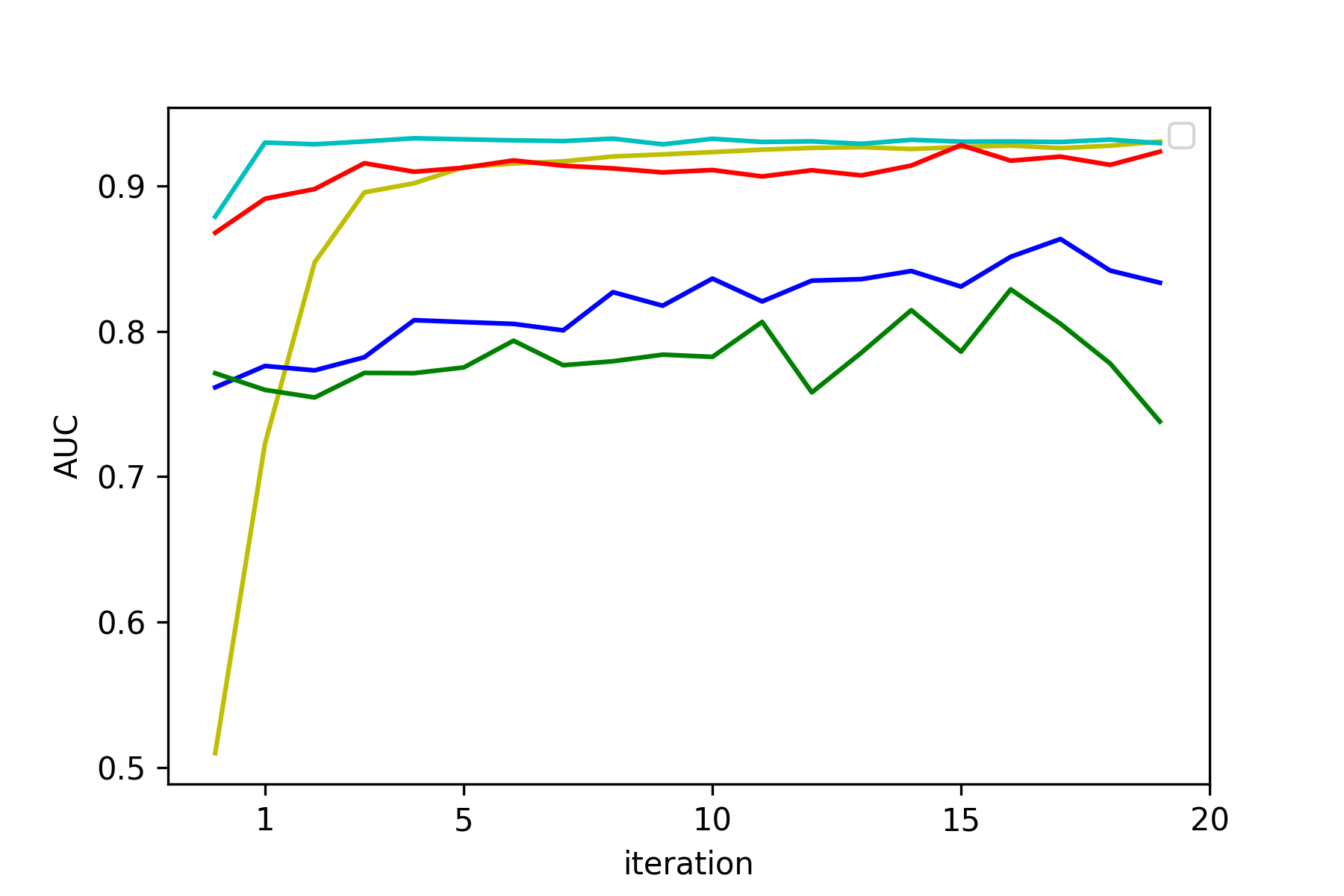}& \hspace{-0.5cm}
\includegraphics[width = 6.5cm,height = 5cm]{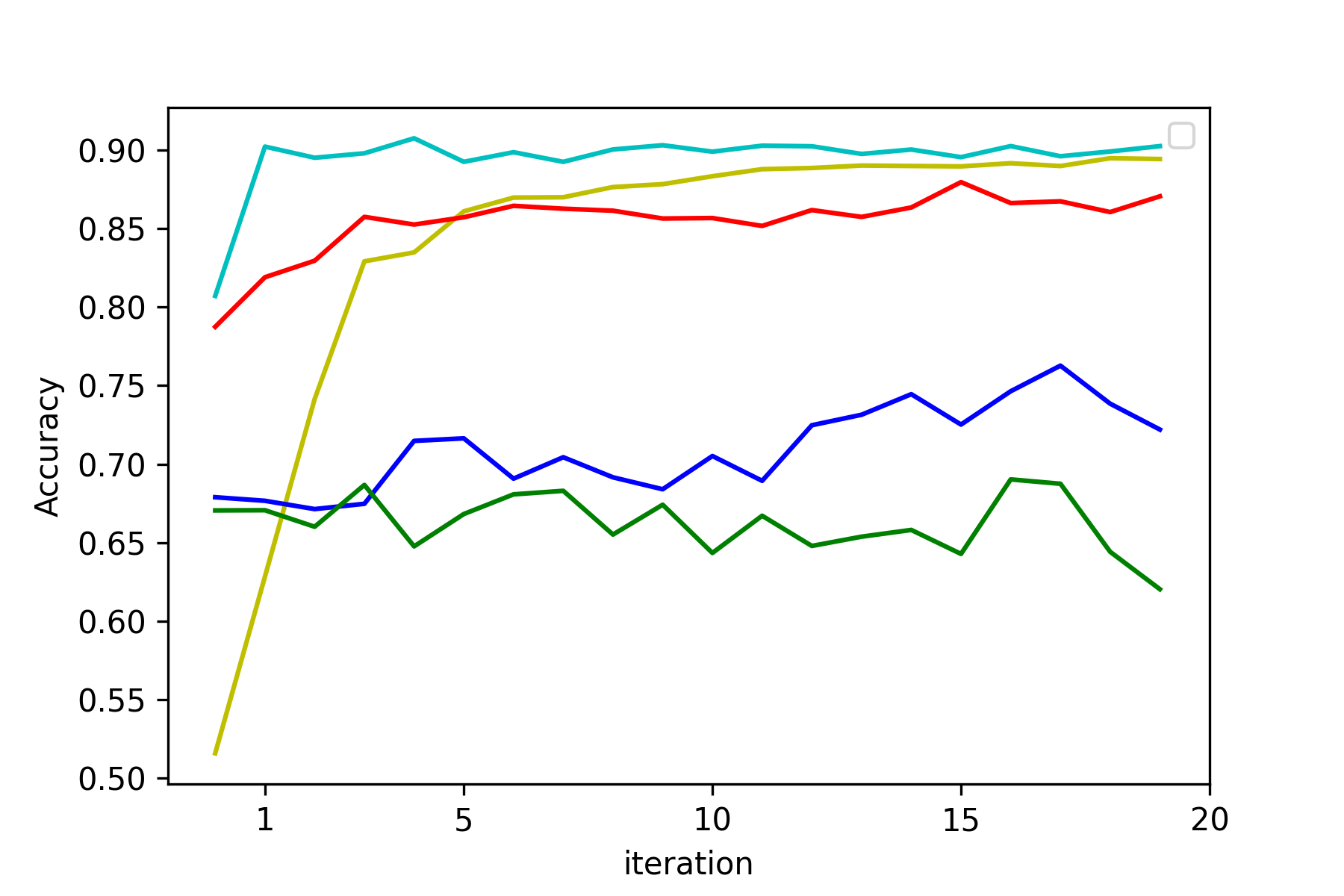}\\
\end{tabular}
\caption{\footnotesize 
\cblue{{Averaged AUC (left) and accuracy (right) on test set along iterations $t \in \{1,\ldots,20\}$ on the control dataset, using PMCnet (yellow), PMCnet-gradfree (blue), PMCnet-covfree (red), DM-PMC (green), and PMCnet-light (cyan).}}}
\label{ROC:syn_AUC_both}
\end{figure}

The results on the test set are summarized in Table~\ref{table:result_syn_diag}. The (probabilistic) ROC curves are provided in Fig.~\ref{ROC:syn_CI_all_diag}, and the metrics evolution (on test set) along training are displayed in Fig.~\ref{ROC:syn_AUC_both}. \cblue{One can notice that PMCnet outperforms its variants PMCnet-gradfree, PMCnet-covfree, and DM-PMC, for all considered metrics. In particular, the gradient adaptation procedure \eqref{eq:AG1} seems necessary to get fast and efficient target exploration (see, for instance, PMCnet vs PMCnet-gradfree curves on Fig.~\ref{ROC:syn_AUC_both}). The covariance adaptation slighlty improves the results in terms of performance metrics. Despite a fast convergence at initial phase (see Fig.~\ref{ROC:syn_AUC_both}), it suffers from a rather high variability (see, for instance, PMCnet vs PMCnet-covfree curves on Fig.~\ref{ROC:syn_CI_all_diag}), and remains slightly below PMCnet and PMCnet-light in terms of classification metrics. The later methods reach very similar metrics, outperforming the competitors. Interestingly, both AUC and accuracy metrics converge much faster using PMCnet-light than its counterpart PMCnet (see Fig.~\ref{ROC:syn_AUC_both})}. Moreover, PMCnet-light displays much less variability, as it can be seen in the std values and the ROC envelopes. This might be an effect of the mini-batch strategy during training phase that allows to update the unknowns after each mini-batch of samples instead of just once for each iteration. It is worth noting that the variability information provided in the ROC envelopes is a mix of the variability inherent to the dataset (which decreases with more training samples) and the variability due to the sampling strategy itself (which decreases with more particles/iterations). %\emilie{Victor, can you help us elaborating on this, as this is quite tricky}

\begin{table}[H]
\hspace{-2cm}
\footnotesize
\cblue{
\begin{tabular}{|c|c|c|c|c|c|c|c|}
\hline
\textbf{Method} & \textbf{AUC} & \textbf{Precision} & \textbf{Recall} & \textbf{Specificity} & \textbf{Accuracy} & \textbf{F1 score} & \textbf{Confusion matrix}\\ 
\hline
{PMCnet} &\textbf{0.9304} (0.0051)&\textbf{0.9410} (0.0196) &\textbf{0.8396} (0.0202)  & \textbf{0.9483} (0.0193) & \textbf{0.8948} (0.0089) &  \textbf{0.8871} (0.0098) &$\begin{bmatrix}
165 & 32\\
10 &193
\end{bmatrix} $ \\
\hline
PMCnet-gradfree & 0.8326 (0.0955) &0.7380 (0.1395) & \textbf{0.8295} (0.1153) & 0.6426 (0.3046)&0.7347 (0.1143) &0.7622 (0.0645)  & $\begin{bmatrix}
163 &34\\
73& 130
\end{bmatrix}$\\
\hline
PMCnet-covfree & {0.9237} (0.0252) &{0.9319} (0.0410) & 0.7988 (0.0431) &{0.9426} (0.0367) &{0.8718} (0.0330) &{0.8596} (0.0366) & $\begin{bmatrix}
157 &40\\
12& 191
\end{bmatrix}$\\
\hline
DM-PMC & 0.7378 (0.1946)&0.6507 (0.1587) &0.7946 (0.2327)  & 0.4512 (0.3903) & 0.6203 (0.1153) &  0.6660 (0.0958) &$\begin{bmatrix}
157 & 40\\
111 &92
\end{bmatrix} $\\
\hline
\hline
{PMCnet-light} &\textbf{0.9336} (0.0009) &\textbf{0.9499} (0.0095) & \textbf{0.8552} (0.0060) &\textbf{0.9561} (0.0093) &\textbf{0.9064} (0.0021) &\textbf{0.9000} (0.0016) & $\begin{bmatrix}
168 &29\\
9& 194
\end{bmatrix}$   \\ 
\hline
\end{tabular}
}
\caption{\small Results on test set for binary classification task of the control dataset with $N_{\text{train}} = 1600$, using {PMCnet} and its variants.}
\label{table:result_syn_diag}
\end{table}

% \begin{table}[H]
% \hspace{-2cm}
% \footnotesize
% \begin{tabular}{|c|c|c|c|c|c|c|c|}
% \hline
% \textbf{Method} & \textbf{AUC} & \textbf{Precision} & \textbf{Recall} & \textbf{Specificity} & \textbf{Accuracy} & \textbf{F1 score} & \textbf{Confusion matrix}\\ 
% \hline
% {PMCnet} &0.9304 (0.0051)&0.9410 (0.0196) &0.8396 (0.0202)  & 0.9483 (0.0193) & 0.8948 (0.0089) &  0.8871 (0.0098) &$\begin{bmatrix}
% 165 & 32\\
% 10 &193
% \end{bmatrix} $ \\
% \hline
% {PMCnet-light} &\textbf{0.9336} (0.0009) &\textbf{0.9499} (0.0095) & \textbf{0.8552} (0.0060) &\textbf{0.9561} (0.0093) &\textbf{0.9064} (0.0021) &\textbf{0.9000} (0.0016) & $\begin{bmatrix}
% 168 &29\\
% 9& 194
% \end{bmatrix}$   \\ 
% \hline
% \end{tabular}
% \caption{\small Results on test set for binary classification task of the control dataset with $N_{\text{train}} = 1600$, using {PMCnet} and {PMCnet-light}, respectively.}
% \label{table:result_syn_diag}
% \end{table}

\subsubsection{Comparison with benchmarks on shallow networks}

We now compare our method with the benchmarks introduced in Sec.~\ref{sec:bench}. We first consider a shallow FCNN with only one hidden layer of few units, i.e., $L=2$ with small $S_{1}$, and three small size classification datasets of LIBSVM library available at \url{https://www.csie.ntu.edu.tw/~cjlin/libsvm/}. On all examples, the dimension of the unknown parameters $d_{\theta}$ is low (few hundreds). 
%Classification and regression tasks are considered, depending on the dataset. Namely the first four datasets are classification problems, while the last one is a regression one. 
The architectures of the chosen shallow FCNNs, as well as the full dataset size (train + validation + test) given by $N_{\text{train}} + 2 N_{\text{test}}$, are summarized in Table~\ref{table:network1} for each dataset. We make use of $tanh$ as the activation function of the hidden layer. 
%As for the dataset \emph{autoMPG}, we choose ReLU. 
We set $(M,K) = (50,100)$ for our {PMCnet} and finetune $T$ for each dataset. {We provide the number of samples per class for the train/validation/test splits of each classification datasets, in Tables~\ref{table:class_ionosphere},~\ref{table:class_wine}, and \ref{table:class_glass}.} We can observe that the dataset \emph{Glass} has a major class imbalance, while the other datasets are relatively balanced. 

\begin{table}[H]
\footnotesize
\hspace{-0.7cm}
\begin{tabular}{|c||c|c|c|c|c|c||c|}
\hline
\textbf{Dataset} & \textbf{Size}  & \textbf{Number of} & \textbf{Number of}  & \textbf{Input} &\textbf{Output} &  \textbf{Number of} &  \textbf{Number of}  \\
& $N_{\text{train}} + 2 N_{\text{test}}$& \textbf{layers $L$} & \textbf{classes $C$} &\textbf{size $S_{0}$} &\textbf{size $d_{y}$} & \textbf{hidden layers $S_{1}$} & \textbf{parameters $d_{\theta}$} \\
\hline
Ionosphere &351& 2 & 2 & 33 &1 & 5& 176\\
\hline
%Diabetes &724 & 2 & 2 &8 &  1 & 3&31 \\
%\hline
Wine & 178&2&3&13&3&3&54\\
\hline
Glass & 214& 2& 6& 9&6&10&166\\
%\hline
%autoMPG & 398 & 2& $\times$ & 9&1&5&56\\
\hline
\end{tabular}
\vspace{0.5cm}
\caption{\small Settings of the shallow FCNN architectures for each dataset. 
%AutoMPG is a regression dataset.
}
\label{table:network1}
\end{table}

\begin{table}[H]
\footnotesize
\centering
\begin{tabular}{|c|c|c|}
\hline
\textbf{Set name} & \textbf{Class 0}  & \textbf{Class 1}  \\ 
\hline
Training set & 81& 129 \\
\hline
Validation set &23 & 47  \\
\hline
Test set & 22&49\\
\hline
\end{tabular}
\vspace{0.5cm}
\caption{\small Label distribution on training, validation, and test sets on dataset \emph{Ionosphere}.}
\label{table:class_ionosphere}
\end{table}

% \begin{table}[H]
% \footnotesize
% \centering
% \begin{tabular}{|c|c|c|}
% \hline
% \textbf{Set name} & \textbf{Class 0}  & \textbf{Class 1}  \\ 
% \hline
% Training set & 279& 155 \\
% \hline
% Validation set &101 & 44  \\
% \hline
% Test set & 95&50\\
% \hline
% \end{tabular}
% \vspace{0.5cm}
% \caption{\small Labels distribution on training, validation and test sets on dataset \emph{Diabetes}.}
% \label{table:class_diabetes}
% \end{table}

\begin{table}[H]
\footnotesize
\centering
\begin{tabular}{|c|c|c|c|}
\hline
\textbf{Set name} & \textbf{Class 1}  & \textbf{Class 2} & \textbf{Class 3} \\ 
\hline
Training set & 31& 44 & 31\\
\hline
Validation set &14 & 13 & 9 \\
\hline
Test set & 14&14&8\\
\hline
\end{tabular}
\vspace{0.5cm}
\caption{\small Label  distribution on training, validation, and test sets on dataset \emph{Wine}.}
\label{table:class_wine}
\end{table}

\begin{table}[H]
\footnotesize
\centering
\begin{tabular}{|c|c|c|c|c|c|c|}
\hline
\textbf{Set name} & \textbf{Class 1}  & \textbf{Class 2} & \textbf{Class 3}& \textbf{Class 4}  & \textbf{Class 5} & \textbf{Class 6} \\ 
\hline
Training set & 45& 42 & 11& 8& 7&15\\
\hline
Validation set &12 & 14 & 4&4&1&8 \\
\hline
Test set & 13&20&2&1&1&6\\
\hline
\end{tabular}
\vspace{0.5cm}
\caption{\small Label  distribution on training, validation, and test sets on dataset \emph{Glass}.}
\label{table:class_glass}
\end{table}

\begin{figure}[H]
\hspace{-1.5cm}
\centering
\includegraphics[width = 9cm,height = 6cm]{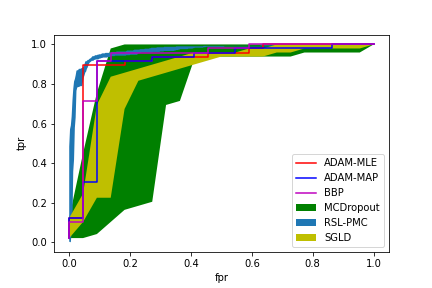}
\caption{\small ROC curves for different estimators, on test set, for dataset \emph{Ionosphere}.}
\label{ROC:binary}
\end{figure}

\begin{figure}[H]
\centering
\begin{tabular}{@{}c@{}c@{}}
\includegraphics[width = 6cm]{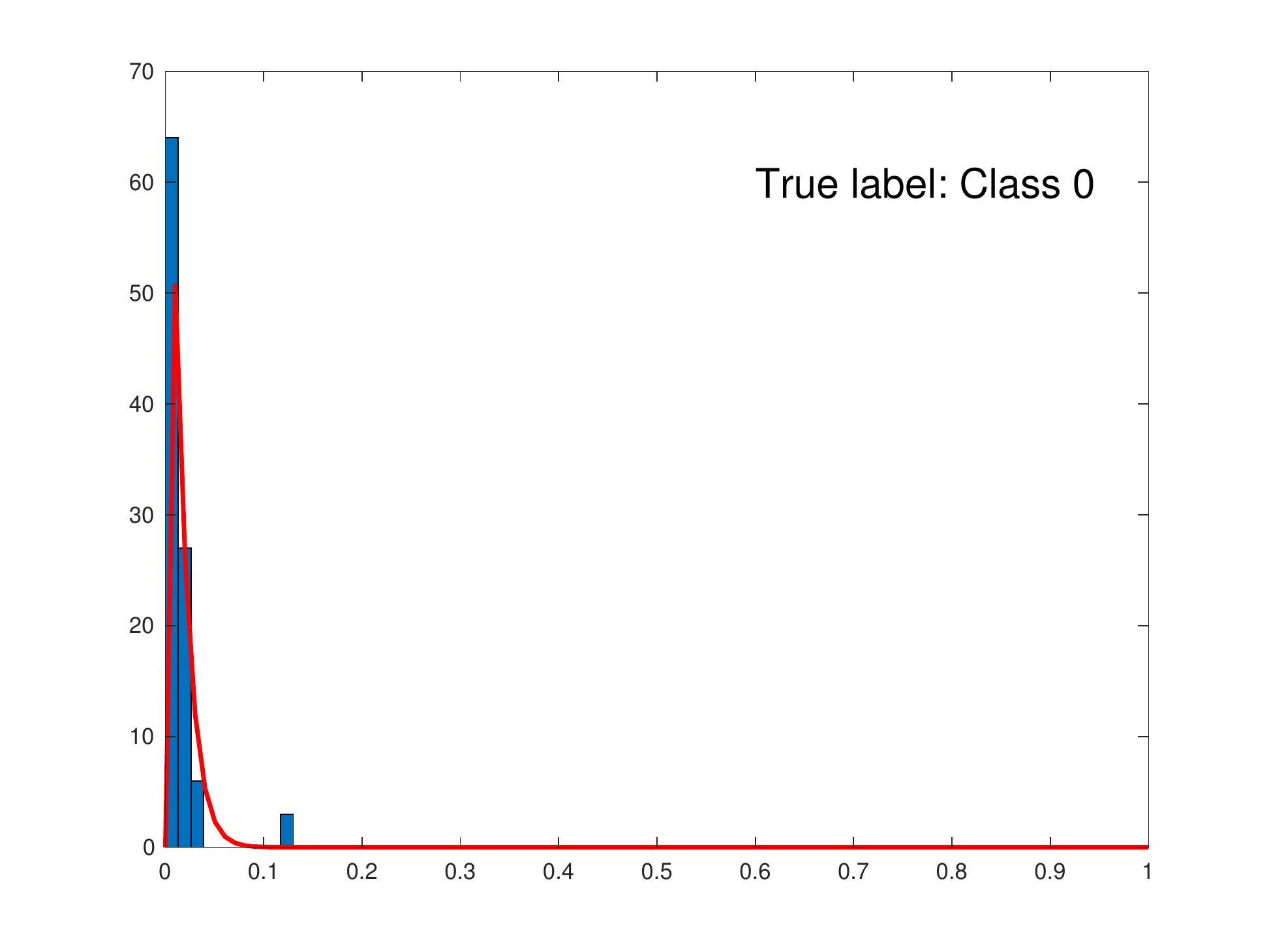}  
&
\includegraphics[width = 6cm]{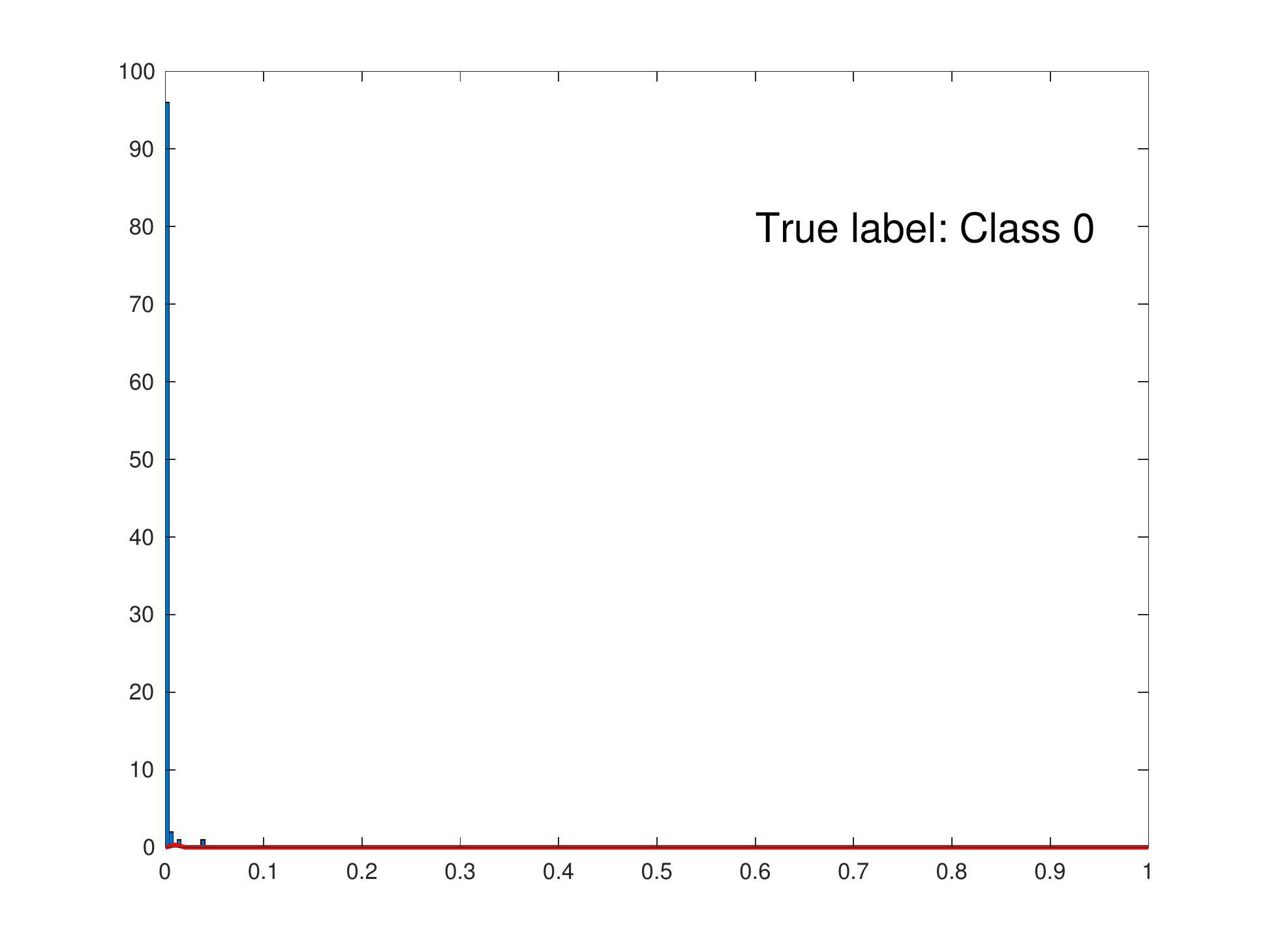} \\
\includegraphics[width = 6cm]{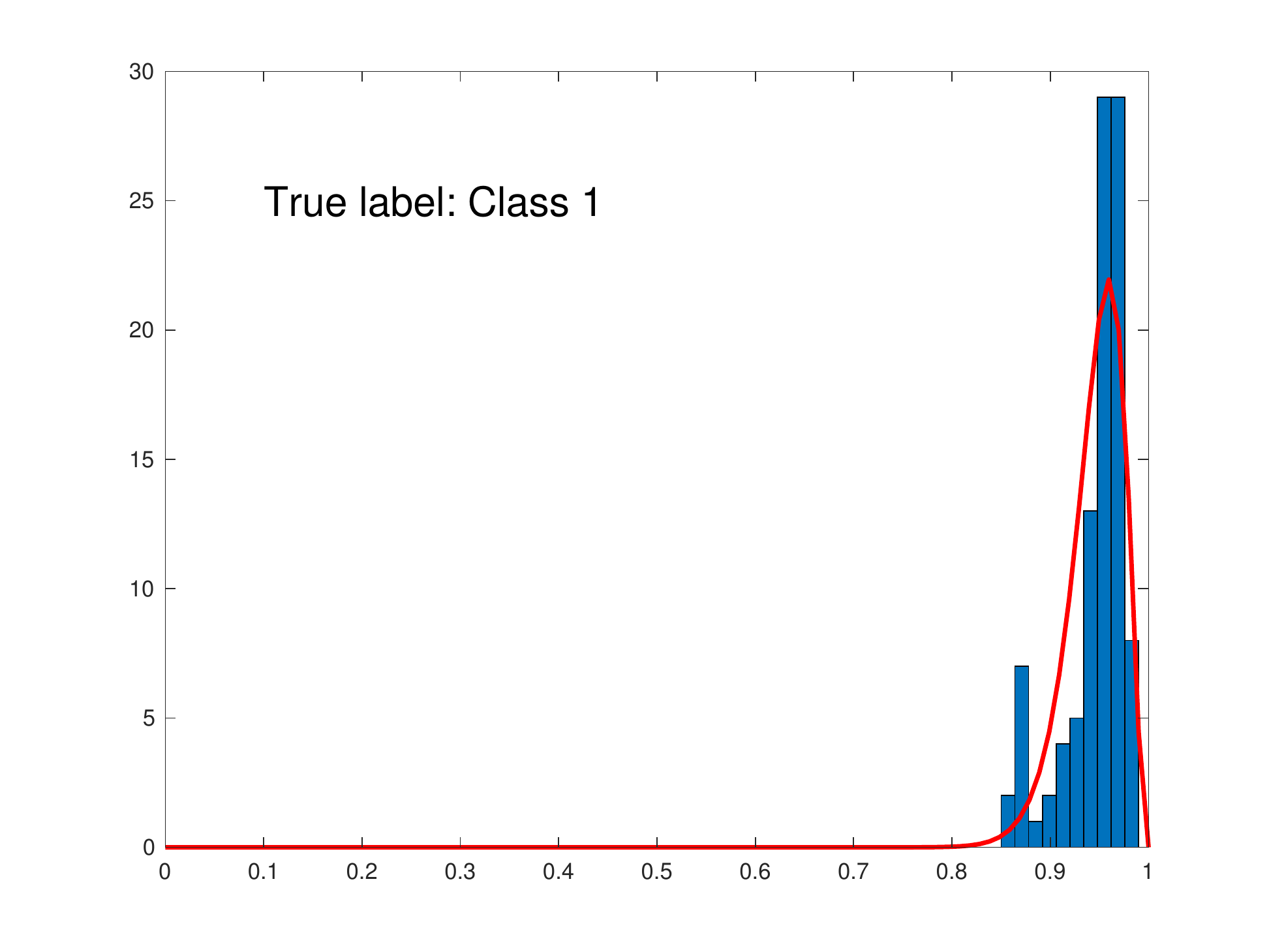}  
&
\includegraphics[width = 6cm]{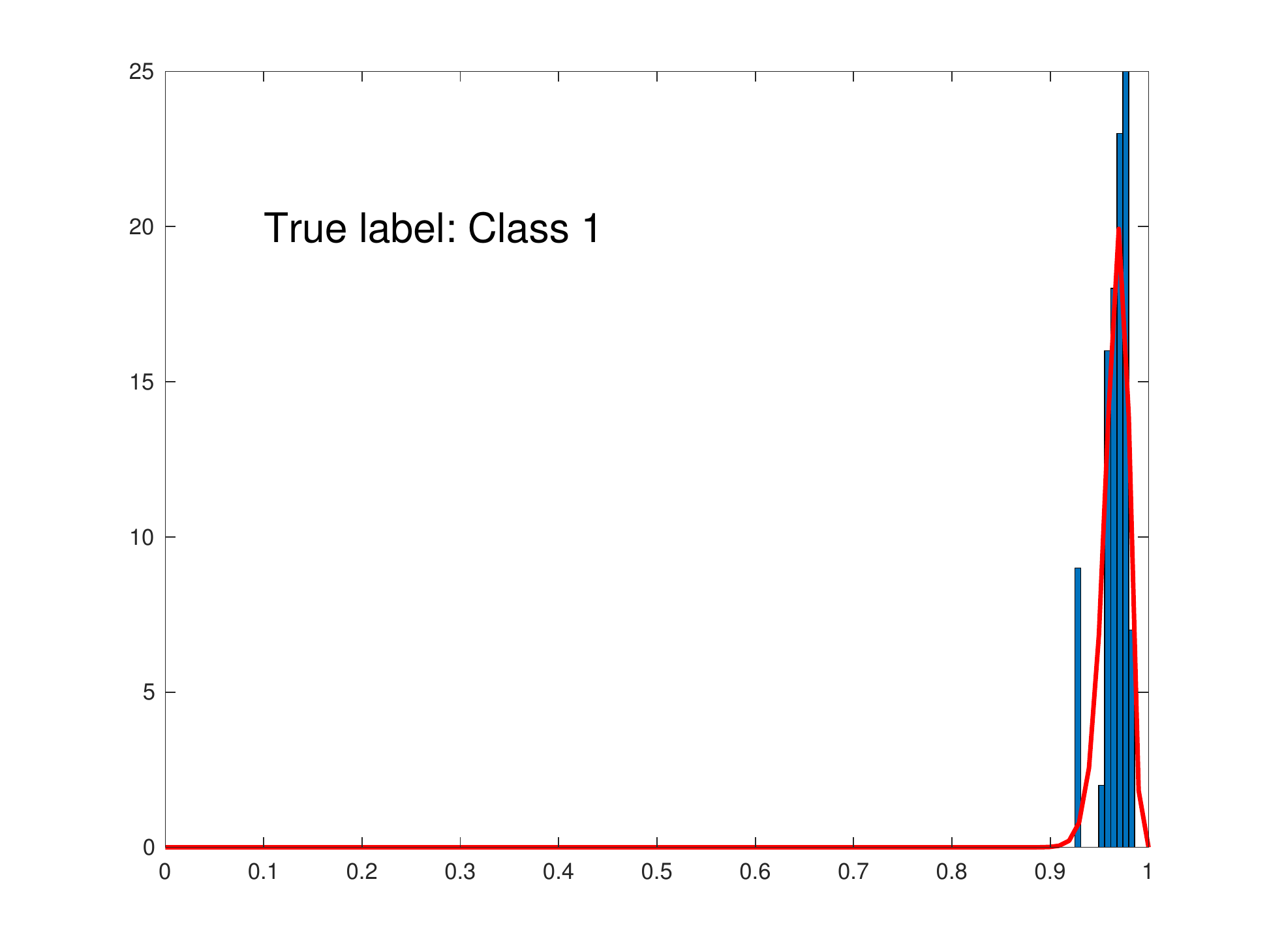} \\
% \includegraphics[width = 6cm]{figures/Diabetes_class0_eg1.pdf}  
% &
% \includegraphics[width = 6cm]{figures/Diabetes_class0_eg2.pdf} \\
% \includegraphics[width = 6cm]{figures/Diabetes_class1_eg1.pdf}  
% &
% \includegraphics[width = 6cm]{figures/Diabetes_class1_eg2.pdf} 
\end{tabular}
\caption{\footnotesize Distributions of the PMCnet network classification decision on four examples extracted from test sets of dataset \emph{Ionosphere}.}
\label{example1}
\end{figure}

Let us start with the results obtained for the binary classification problem with dataset \emph{Ionosphere}. We summarize the metrics obtained on test sets in Table~\ref{table:result1}. Our finetuning led to $T=50$. We can observe that  our proposed method reaches best precision, specificity, accuracy, and F1 score among all the methods, with a small variability. ADAM-MAP reaches good performance too on this experiment. Among Bayesian-based competitors, MCDropout, and SAE reach the best results, however with lower accuracy than our method. We also display ROC curves (and CIs, when available) in Fig.~\ref{ROC:binary}. This illustrates again that our proposed algorithm provides a good predictive performance with small variability. In Fig.~\ref{example1}, we pick some examples from the test sets of the dataset and display the histograms (with $10$ bins) of $\{\y^{(r,n)}\}_{1 \leq r \leq R}$ obtained by Algorithm~\ref{tab:PMCnetoutputmetric}. For better readability, we also superimpose a red curve obtained by simply fitting a beta distribution on the obtained histograms. One can see that our proposed algorithm has the ability to provide a meaningful probabilistic information about its binary classification decision. For instance, for the example displayed in top right of Fig.~\ref{example1}, our method gives the classification decision with high confidence, leading to peaky histogram. In contrast, on the other examples, the method shows more variability (i.e., less confidence) in its decision, leading to a more spread histogram for the estimated mean. Such information can be of high interest for practitioners, in sensitive fields such as healthcare.

\begin{table}[H]
\scalebox{0.8}{
\hspace{-2.5cm}
\begin{tabular}{|c|c|c|c|c|c|c|c|}
\hline
\textbf{Method} & \textbf{AUC} & \textbf{Precision} & \textbf{Recall} & \textbf{Specificity} & \textbf{Accuracy} & \textbf{F1 score} & \textbf{Confusion matrix}\\ 
\hline
ADAM-MLE &0.9796 &0.8077 &\textbf{0.9545} & 0.8980&0.9155 &0.8750  & $\begin{bmatrix}
21 &1\\
5& 44
\end{bmatrix}$   \\ 
\hline
ADAM-MAP &\textbf{0.9814} &0.8333 &0.9091 & 0.9184&0.9155 &0.8696  & $\begin{bmatrix}
20 &2\\
4& 45
\end{bmatrix}$   \\ 
\hline
% BNN-last2&0.9617 (0.0139)& 0.8473 (0.0665)&0.9094 (0.0235) &0.9234 (0.0383)  &0.9190 (0.0270) & 0.8757 (0.0372)&$\begin{bmatrix}
%20 &2\\
%4& 45
%\end{bmatrix}$\\
%\hline
% BNN-secondhalf&0.9618 (0.0139)& 0.8464 (0.0663)&0.9095 (0.0235) &0.9229 (0.0383)  &0.9187 (0.0270) & 0.8753 (0.0371)&$\begin{bmatrix}
%20 &2\\
%4& 45
%\end{bmatrix}$\\
%\hline
 BBP&0.9137 &0.8333 &0.9091  & 0.9184 & 0.9155 &  0.8696&$\begin{bmatrix}
20 & 2\\
4 &45
\end{bmatrix} $ \\ 
\hline
SGLD&0.8824 (0.0110)& 0.6959 (0.0132) &0.8100 (0.0175)  &0.8410 (0.0083)  & 0.8314 (0.0088) & 0.7486 (0.0134) &$\begin{bmatrix}
18 &4\\
8 &41
\end{bmatrix} $ \\ 
\hline
MCDropout&0.9159 (0.0186) &0.8678 (0.0349)  &0.7736 (0.0514)  & \textbf{0.9465} (0.0160) & 0.8930 (0.0177) &  0.8169 (0.0335)&$\begin{bmatrix}
17 &5\\
3 & 46
\end{bmatrix} $ \\ 
\hline
SAE& 0.9064 (0.0036)  & 0.8268 (0.0272)  & 0.8609 (0.0108)  &  0.9186 (0.0151)  & 0.9007 (0.0120) & 0.8433 (0.0170) &$\begin{bmatrix}
19 & 3\\
4 & 45
\end{bmatrix} $ \\ 
\hline
{PMCnet}&0.9642 (0.0146)& \textbf{0.8543} (0.0684)&0.9145 (0.0216) &0.9269 (0.0391)  &\textbf{0.9231} (0.0283) & \textbf{0.8819} (0.0393)&$\begin{bmatrix}
20 &2\\
4& 45
\end{bmatrix}$ \\
\hline
\end{tabular}
}
\caption{\small Results for binary classification, computed on test set, on dataset \emph{Ionosphere}.}
\label{table:result1}
\end{table}

\begin{figure}[H]
\scalebox{0.8}{
\hspace{-5cm}
\begin{tabular}{@{}c@{}c@{}c@{}c@{}c@{}c@{}c@{}}
\includegraphics[width = 4cm]{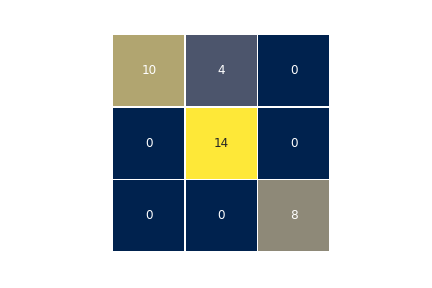}\hspace{-1.5cm}& \hspace{-1cm}
\includegraphics[width = 4cm]{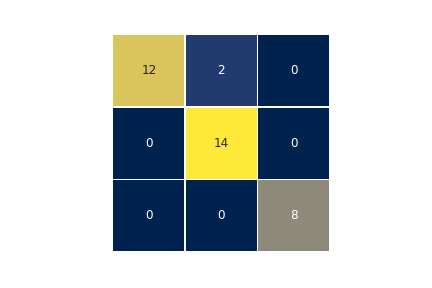}\hspace{-1.5cm}& \hspace{-1cm}
\includegraphics[width = 4cm]{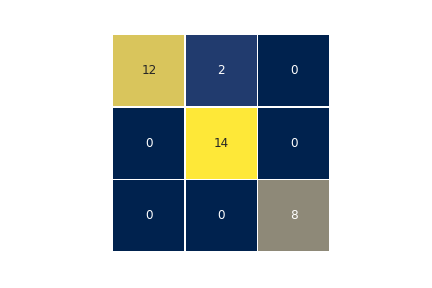} \hspace{-1.5cm}& \hspace{-1cm}
\includegraphics[width = 4cm]{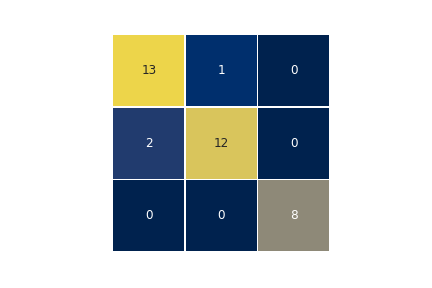}\hspace{-1.5cm}& \hspace{-1cm}
\includegraphics[width = 4cm]{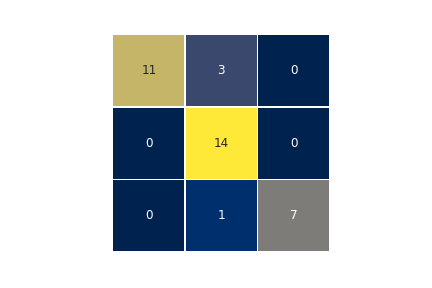}\hspace{-1.5cm}& \hspace{-1cm}
\includegraphics[width = 4cm]{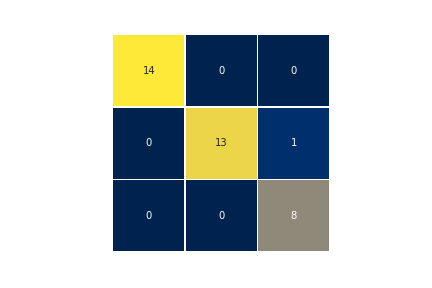}\hspace{-1.5cm}& \hspace{-1cm}
\includegraphics[width = 4cm]{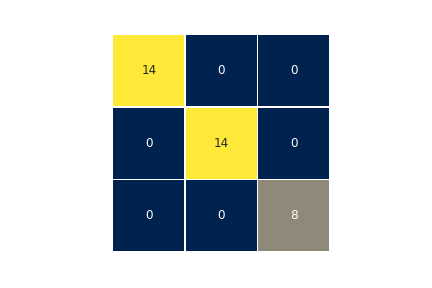} \\
ADAM-MLE & \hspace{-0.5cm}ADAM-MAP &\hspace{-0.5cm} BBP &  \hspace{-0.5cm}SGLD & \hspace{-0.5cm} MCDropout &  \hspace{-0.5cm} SAE & \hspace{-0.5cm}{PMCnet} \\
\includegraphics[width = 4cm]{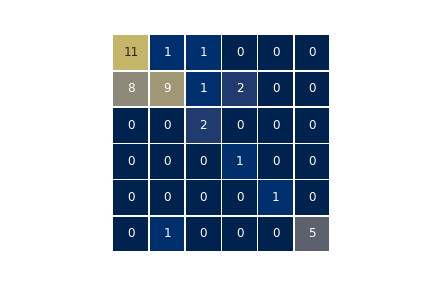}\hspace{-1.5cm}& \hspace{-1cm}
\includegraphics[width = 4cm]{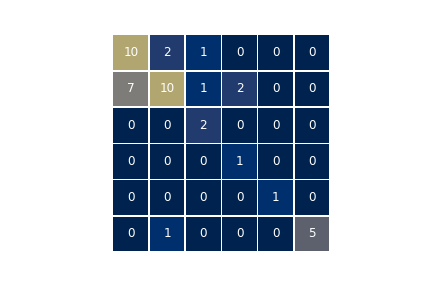}\hspace{-1.5cm}& \hspace{-1cm}
\includegraphics[width = 4cm]{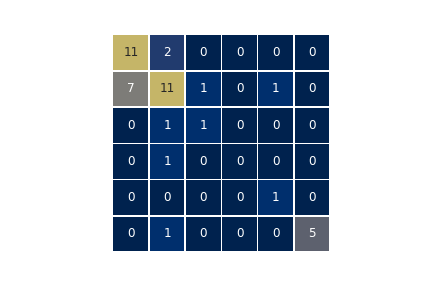} \hspace{-1.5cm}& \hspace{-1cm}
\includegraphics[width = 4cm]{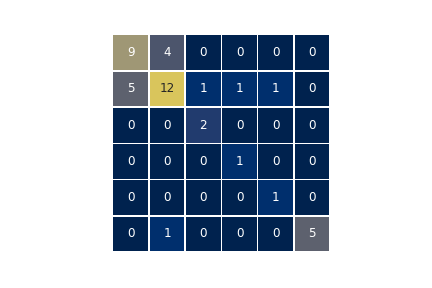}\hspace{-1.5cm}& \hspace{-1cm}
\includegraphics[width =
4cm]{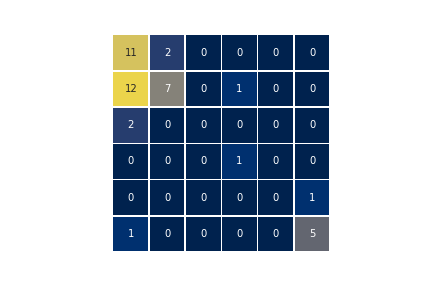}\hspace{-1.5cm}& \hspace{-1cm}
\includegraphics[width =
4cm]{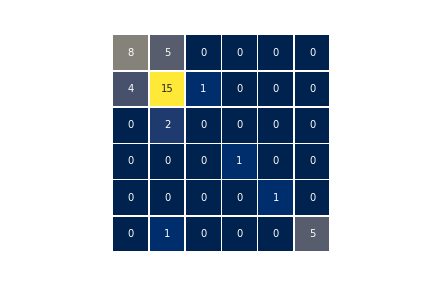}\hspace{-1.5cm}& \hspace{-1cm}
\includegraphics[width = 4cm]{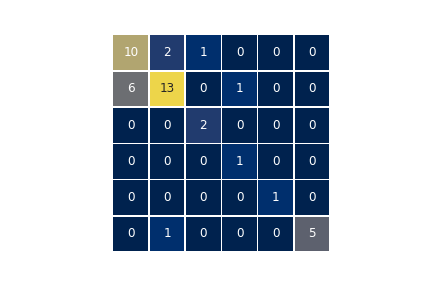} \\
ADAM-MLE &  \hspace{-0.5cm}ADAM-MAP &  \hspace{-0.5cm}BBP &  \hspace{-0.5cm}SGLD &  \hspace{-0.5cm}MCDropout &\hspace{-0.5cm}SAE &  \hspace{-0.5cm}{PMCnet} \\
\end{tabular}
}
\caption{\footnotesize Confusion matrices on test set, for the different benchmarks on dataset \emph{Wine} (top) and dataset \emph{Glass} (bottom).}
\label{confusion1}
\end{figure}

\begin{figure}[H]
\centering
\begin{tabular}{@{}c@{}c@{}}
\includegraphics[width = 6.2cm]{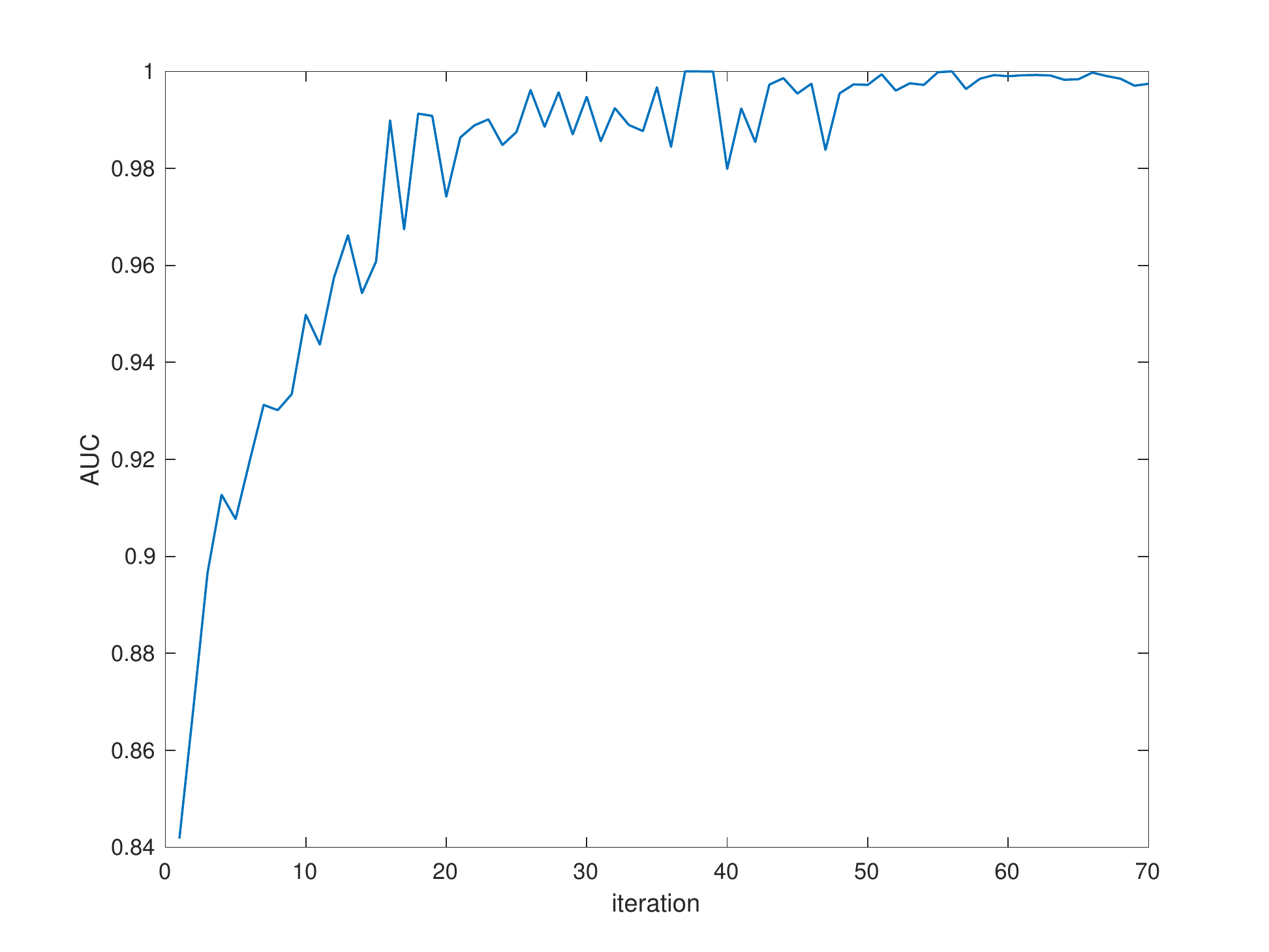}  
&
\includegraphics[width = 6.2cm]{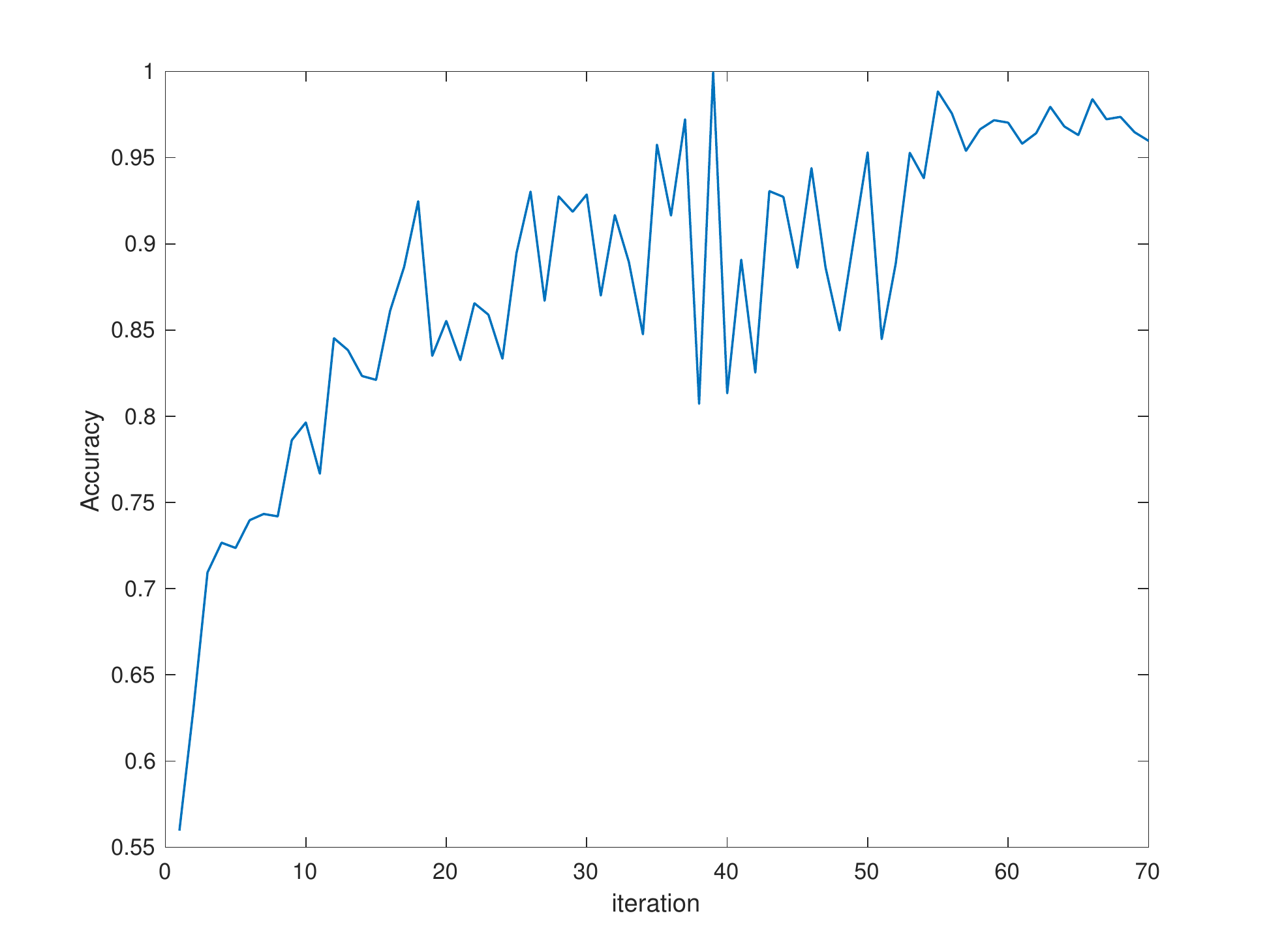} 
\end{tabular}
\caption{\footnotesize Averaged AUC and accuracy curves, on test set, as a function of iterations during training using {PMCnet}, on dataset \emph{Wine}.}
\label{AUC:wine}
\end{figure}

Next, we focus on the multi-class classification problems related to datasets \emph{Wine} and \emph{Glass}. The numerical results are summarized in Table~\ref{table:result2}. We can see again that our methods gives the best accuracy for both datasets. Even more, for the dataset \emph{Wine}, it can perfectly predict the class of wine on the considered test set. As for the dataset \emph{Glass}, the performance of every method is not as good as those for the dataset \emph{Wine}, probably due to the high class imbalance in this dataset. Our method still outperforms others in this dataset with acceptable variability. We also draw the heat map of the confusion matrix associated to these results in Fig.~\ref{confusion1}. Clearly, the PMCnet method leads to confusion matrices with the most zeros on the off-diagonal axis.
%we provide the ROC curves (and CIs, when available) for each class in Fig.~\ref{ROC:wine}. We also
For dataset \emph{Wine}, we plot the mean of AUC and accuracy of test set along iterations during the training phase of {PMCnet} in Fig.~\ref{AUC:wine}. We can see that, as the training goes on, both AUC and accuracy tend to stabilize, which shows the validity of our method. %Hereagain, we build distribution plots for the PMCnet classification decision using procedure in Algorithm~\ref{tab:PMCnetoutput}. We provide such plots for two examples per class picked on the test set of dataset \emph{Wine} in Fig.~\ref{example:wine}. We can see that in these examples, the variability on the classification decision is rather low, showing a high confidence of the method in this dataset.

\begin{table}[H]
\centering
\footnotesize
\begin{tabular}{|c|c|c|c|c|}
\hline
&\textbf{Method} & \textbf{AUC} & \textbf{F1 score} & \textbf{Accuracy}\\
\hline
  \multirow{6}{*}{\STAB{\rotatebox[origin=c]{90}{Wine}}}&ADAM-MLE &0.9950 &0.9028 &0.8889\\
\cline{2-5}
&ADAM-MAP &\textbf{0.9999} &0.9521 &0.9286\\
\cline{2-5}
%BNN-last2 &\textbf{1.0000} (0) &0.9914 (0.0183)&  $\begin{bmatrix}
%14 &0 & 0\\
%0& 14& 0\\
%0& 0 & 8\\
%\end{bmatrix}$  \\ 
%\hline
%BNN-secondhalf &\textbf{1.0000} (0.0001) &0.9954 (0.0105)&  $\begin{bmatrix}
%14 &0 & 0\\
%0& 14& 0\\
%0& 0 & 8\\
%\end{bmatrix}$  \\ 
%\hline
&BBP &0.9912 & 0.9761 & 0.9444\\
\cline{2-5}
&SGLD &0.9776 (0.0046)& 0.9396 (0.0202) &0.9297 (0.0234)\\
\cline{2-5}
&MCDropout &0.9857 (0.0087) &0.8929 (0.0415) & 0.8867 (0.0402)\\
\cline{2-5}
&SAE & 0.9938 (0.0151)& 0.9669 (0.0609) & 0.9836 (0.0290) \\
\cline{2-5}
&{PMCnet} &\textbf{0.9974} (0.0048) & \textbf{0.9951} (0.0191) &\textbf{0.9944} (0.0215)\\
\hline
  \multirow{6}{*}{\STAB{\rotatebox[origin=c]{90}{Glass}}}&ADAM-MLE &0.8361&0.7240 &0.6742\\
\cline{2-5}
&ADAM-MAP &0.8347& 0.7247 &0.6744\\
\cline{2-5}
% BNN-last2&\textbf{0.7904} (0.0137)& \textbf{0.7670} (0.0184)&0.8229 (0.0316) &\textbf{0.5233} (0.0564)  &0.7196 (0.0182) & 0.7935 (0.0149)&$\begin{bmatrix}
%78 &17\\
%24& 26
%\end{bmatrix}$\\
%\hline
% BNN-secondhalf&0.7885 (0.0164)& 0.7628 (0.0184)&0.8254 (0.0314) &0.5105 (0.0585)  &0.7168 (0.0187) & 0.7924 (0.0150)&$\begin{bmatrix}
%78 &17\\
%24& 26
%\end{bmatrix}$\\
%\hline
& BBP&0.8360 &0.6187 &0.6512 \\
\cline{2-5}
&SGLD&0.8498 (0.0029) & 0.7259 (0.0105) &0.7026 (0.0095)  \\
\cline{2-5}
&MCDropout&0.7877 (0.0297)&0.4045 (0.0504) & 0.5502 (0.0486) \\
\cline{2-5}
&SAE&0.7680 (0.0053)& 0.7087 (0.1499)& 0.7151 (0.0128) \\
\cline{2-5}
&{PMCnet}&\textbf{0.8510} (0.0059)& \textbf{0.7715} (0.0580) & \textbf{0.7414} (0.0264)\\
\hline
\end{tabular}
\caption{\small Results for multi-class classification on dataset \emph{Wine} and dataset \emph{Glass} respectively.}
\label{table:result2}
\end{table}

% \begin{figure}[H]
% \hspace{-3cm}
% \begin{tabular}{@{}c@{}c@{}c@{}}
% \includegraphics[width = 6.5cm,height = 5cm]{figures/ROC_Wine_BNN_class1_new.pdf}& \hspace{-0.5cm}
% \includegraphics[width = 6.5cm,height = 5cm]{figures/ROC_Wine_BNN_class2_new.pdf}& \hspace{-0.5cm}
% \includegraphics[width = 6.5cm,height = 5cm]{figures/ROC_Wine_BNN_class3_new.pdf}  \\
% (a) & (b) & (c)\\
% \end{tabular}
% \caption{\footnotesize ROC curves for {PMCnet}, on test set, for dataset \emph{Wine}, on a one-versus-all decision associated to (a) Class 1, (b) Class 2 and (c) Class 3 respectively.}
% \label{ROC:wine}
% \end{figure}

\begin{figure}[H]
\centering
\begin{tabular}{@{}c@{}c@{}}
\includegraphics[width = 6.2cm]{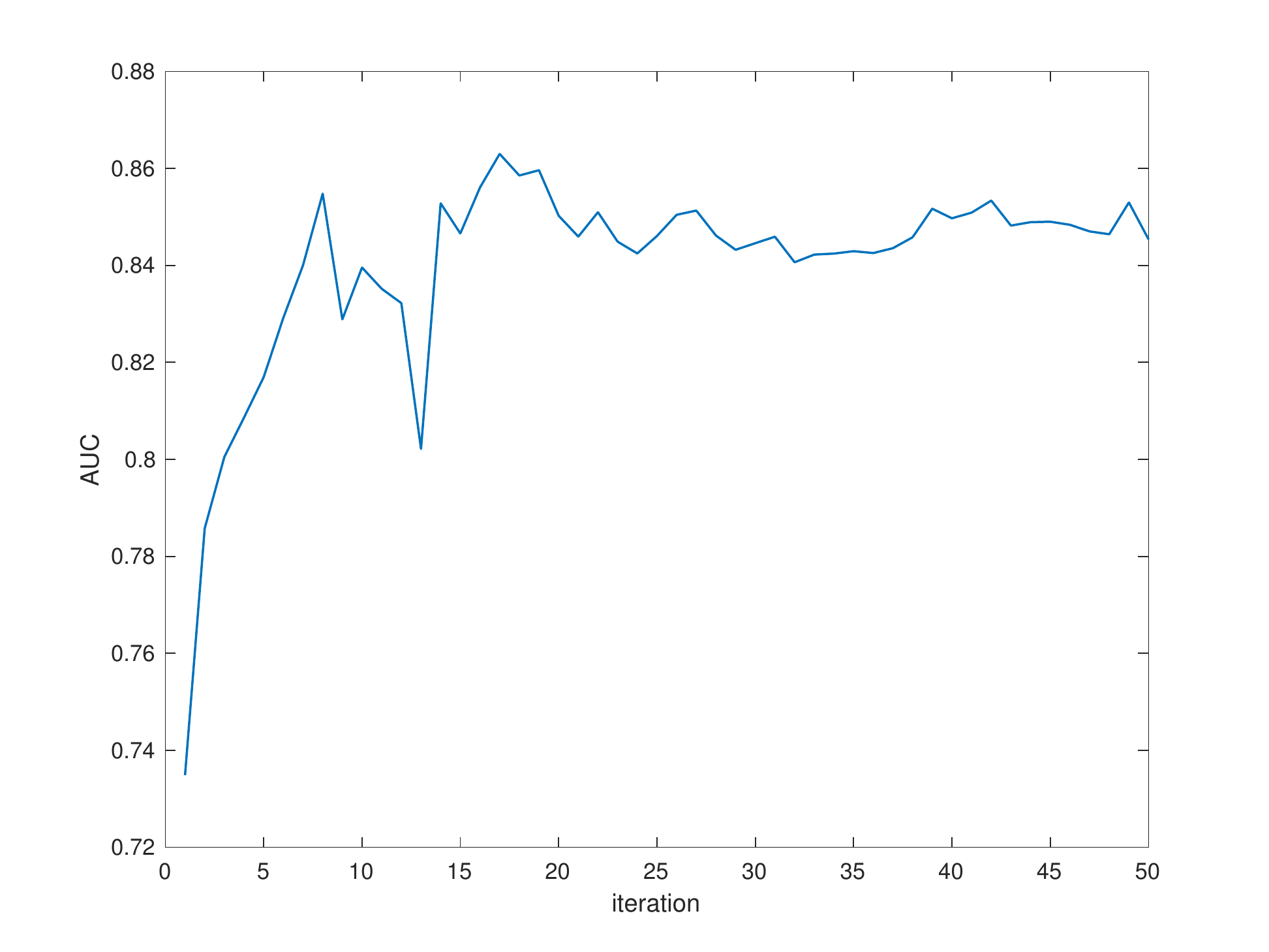}  
&
\includegraphics[width = 6.2cm]{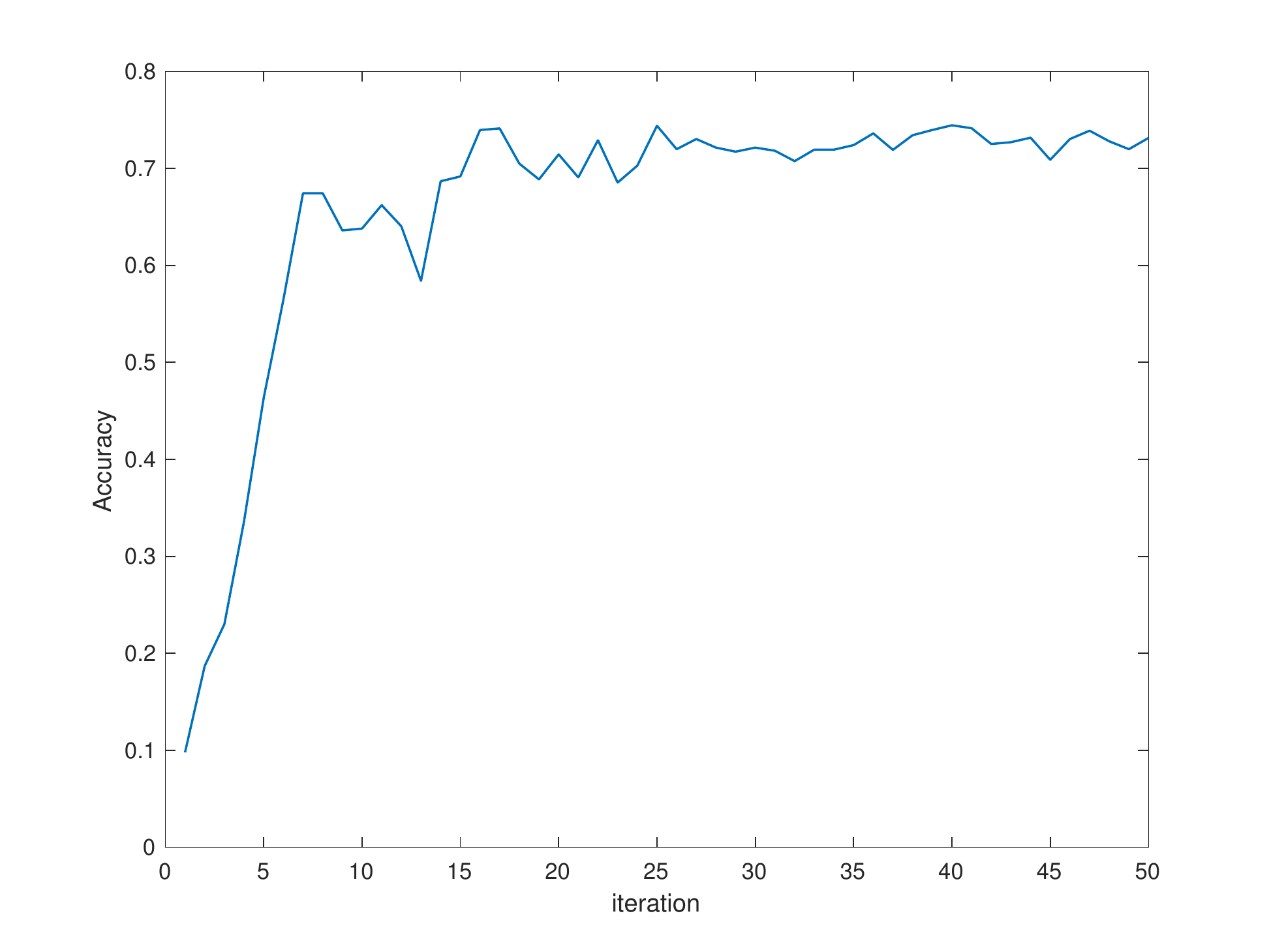} 
\end{tabular}
\caption{\footnotesize Averaged AUC and accuracy curves, on test set, as a function of iterations during training using {PMCnet}, on dataset \emph{Glass}.}
\label{AUC:glass}
\end{figure}

As for dataset \emph{Glass}, we plot the mean of AUC and accuracy of test set along iterations in Fig.~\ref{AUC:glass}. Again, we can see the proposed method converges gradually. %In Fig.~\ref{ROC:glass}, we plot the ROC curves for each class on dataset \emph{Glass}, we can see the good performance of our method with small confidence interval for different classes. 
%For each class, we choose one example to display the distribution of the classification decision  given by our method in Fig.~\ref{example:glass}.
%We can see that, compared to dataset \emph{Wine}, the classifier generated by our method is less accurate for dataset \emph{Glass}. Especially for Class 5, some misclassified decision can happen for some of the samples from the approximated posterior. The interest of our method is to allow the access to the full distribution of the classification decision, instead of only a point-wise one, yielding a very useful assessment for the practitioner. 

\subsubsection{Deep networks}
We now move to comparisons of the methods for learning the parameters of DNNs, from larger scale datasets. Namely, we focus on \cblue{\emph{MNIST-C} \cite{mu2019mnist}, \emph{SVHN-G}\footnote{\url{http://ufldl.stanford.edu/housenumbers/}} and \emph{CIFAR10}\footnote{\url{https://www.cs.toronto.edu/~kriz/cifar.html}} datasets} for multi-class classification task. \cblue{For the regression task, we evaluate \emph{Protein}\footnote{\url{https://archive.ics.uci.edu/ml/datasets/Physicochemical+Properties+of+Protein+Tertiary+Structure}} and \emph{Naval}\footnote{\url{https://www.kaggle.com/datasets/elikplim/maintenance-of-naval-propulsion-plants-data-set}}}. Note that in the original dataset \emph{MNIST-C}, there are several different variants of the input images, here we are only using hand-written digits possibly corrupted by rotation. The sample size for training set, validation set, and test set is 45000, 15000, and 10000 respectively, involving grayscale digit images of size $28\times 28$ as input. \cblue{For dataset \emph{SVHN-G}, the sample size for training set, validation set, and test set is 54942, 18315, and 26032 respectively, involving color digit images of size $32\times 32$ as input. For dataset \emph{CIFAR10}, the sample size for training set, validation set, and test set is 45000, 5000, and 10000 respectively, involving color natural images of size $32\times 32$ as input.} \cblue{In \emph{MNIST-C}, \emph{SVHN-G}, and \emph{CIFAR10} datasets, there are $10$ classes to distinguish}. The architectures for the DNNs employed for the datasets are described in Table~\ref{table:network2}. \cblue{For datasets \emph{MNIST-C} and \emph{SVHN-G}}, we retain LeNet-5 as the DNN for the classifier as it was used in 2021 NeurIPS competition \cite{Andrew2021}. \cblue{For dataset \emph{CIFAR10}, we retain ResNet-20 as the DNN for the classifier. {As for datasets \emph{Protein} and \emph{Naval}, we adopt an FCNN with two wide hidden layers of $100$ units (i.e., $L = 3$, $S_1 = S_2 = 100$) and we choose ReLU as the activation function for hidden layers.}} On such large datasets, only {PMCnet-light} version can be used, and we set $(M,K) = (20,20)$ with $T$ finetuned for each dataset.

\begin{table}[H]
\footnotesize
\centering
\cblue{
\begin{tabular}{|c||c|c|c|c|c||c|}
\hline
\textbf{Dataset} & \textbf{Sample}  & \textbf{Networks} & \textbf{Number of}  & \textbf{Input} &\textbf{Output}  &  \textbf{Number of}  \\
&\textbf{size}&& \textbf{classes $C$} &\textbf{size $S_{0}$} &\textbf{size $d_{y}$} & \textbf{parameters $d_{\theta}$} \\
\hline
MNIST-C  &70000& LeNet-5 & 10 &784  &10 & 61706\\
\hline
SVHN-G & 99289 & LeNet-5 & 10 &1024  &10 & 82826\\
\hline
CIFAR10 & 60000 & ResNet-20 & 10 & 1024  &10 & 269722 \\
\hline
Protein & 45730 & FCNN & $\times$ &9 &  1 &11201 \\
\hline
Naval & 11934 & FCNN & $\times$ &  14 &  1 & 11701  \\
\hline
\end{tabular}
}
\vspace{0.5cm}
\caption{\small Settings of the DNN architectures retained for classification and regression datasets.}
\label{table:network2}
\end{table}

\begin{figure}[H]
\begin{tabular}{@{}c@{}c@{}c@{}c@{}}
\hspace{-5cm}
\includegraphics[width = 7cm]{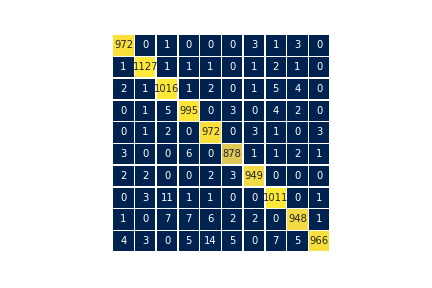}\hspace{-2cm}&\hspace{-2cm}
\includegraphics[width = 7cm]{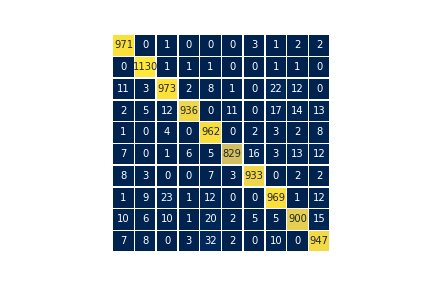}\hspace{-2cm}& \hspace{-2cm}
\includegraphics[width = 7cm]{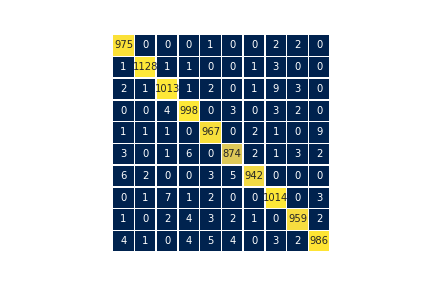}\hspace{-2cm}&\hspace{-2cm}
\includegraphics[width = 7cm]{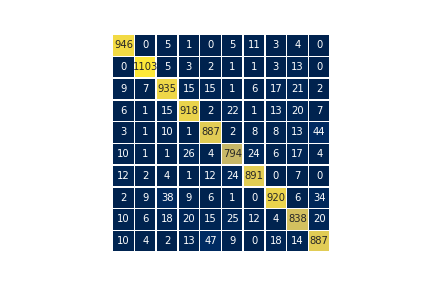}\hspace{-2cm}\vspace{-0.5cm} \\
\hspace{-5cm} MLE &\hspace{-2cm} MAP & \hspace{-2cm}BBP &\hspace{0.2cm} SGLD\\
\hspace{-5cm}
\includegraphics[width = 7cm]{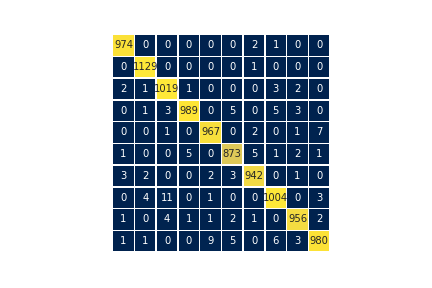}\hspace{-2cm}&\hspace{-2cm}
\includegraphics[width = 7cm]{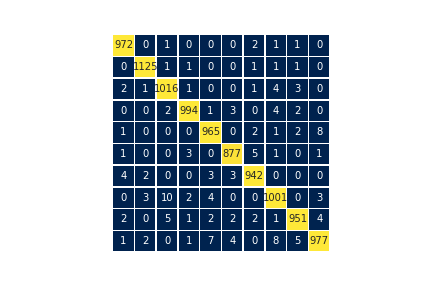}\hspace{-2cm}& \hspace{-2cm}
\includegraphics[width = 7cm]{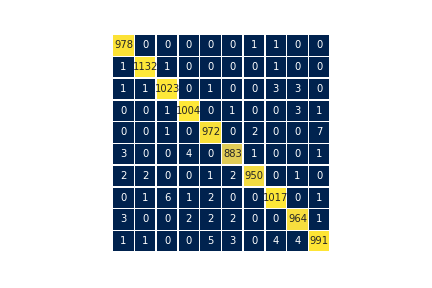}\vspace{-0.5cm} & \\
\hspace{-5cm}MCDropout & \hspace{-2cm}SAE &\hspace{-2cm} {PMCnet-light} & \\
\end{tabular}
\vspace{-0.2cm}
\caption{\footnotesize Confusion matrices for the different benchmarks on test set of dataset \emph{MNIST-C}.}
\label{matrix:MNIST_C}
\end{figure}

\cblue{First, we provide the results obtained for different methods on the classification datasets in Table~\ref{table:result6} by choosing $T=20$. Hereagain, the classifier derived from our method is the best among all the competitors in terms of both AUC and accuracy, for the three datasets. Note that, on these datasets, almost all the probabilistic methods make predictions with small variability. Even so, our proposed method still performs best with a mean AUC close to 1 and high F1 score.
%, nearly one percent higher than benchmarks. 
Considering the sample size of these large datasets, the improvement with respect to the competitors is significant. 
%Our methods also reaches the best averaged F1 score. 
The heat map for confusion matrix of each method for \emph{MNIST-C} dataset is provided in Fig.~\ref{matrix:MNIST_C}, we can see that for each class, {PMCnet-light} is able to predict most digits correctly.}

% Misclassification usually happens when the classifier recognizes digit 4 as digit 9. To visualize the results better, we also choose four typical examples (with correct or incorrect decision) derived by {PMCnet-light} in Fig.~\ref{example:MNIST_C}. In the two first examples, one can see that the classifier identifies well the digit even if it is rotated. The two last examples show failure of the method. We can see that these examples are even difficult for humans to recognize especially due to the rotation and incomplete digit writing. 

% We also provide in Fig.~\ref{hist:MNIST_C} the averaged network outputs (of dimension $10$ as the number of classes) predicted by our method for these four examples. We can see that for the first examples, {PMCnet-light} predicts the digit correctly with maximal probability. However, for the two last examples, our method shows two different behaviours. On the third case, it is very certain to detect (wrongly) digit $4$. While for the fourth example, the method gives non-zero probability for digits $1$, $6$ and $8$.

\begin{table}[H]
\hspace*{-1.2cm}
\footnotesize
\centering
\cblue{
\begin{tabular}{|c|c|c|c|c|}
\cline{2-5}
\multicolumn{1}{c|}{ } & \textbf{Method} & \textbf{AUC}& \textbf{F1 score} &\textbf{Accuracy} \\ 
\hline
\parbox[t]{2mm}{\multirow{7}{*}{\rotatebox[origin=c]{90}{Dataset \emph{MNIST-C}}}} & ADAM-MLE&0.9998&0.9833&0.9834    \\ 
\cline{2-5}
& ADAM-MAP&0.9988&0.9547&0.9550    \\ 
\cline{2-5}
& BBP&0.9998 &0.9855 &0.9856   \\ 
\cline{2-5}
& SGLD & 0.9949 (0.0008) & 0.9115 (0.0135)&0.9126 (0.0132)  \\ 
\cline{2-5}
& MCDropout & 0.9998 (0) &0.9837 (0.0008) &0.9838 (0.0008) \\ 
\cline{2-5}
& SAE & 0.9998 (0) &0.9824 (0.0009) &\textbf{0.9998} (0.0003) \\ 
\cline{2-5}
& {PMCnet-light}& \textbf{1.0000} (0)&\textbf{0.9913} (0.0002) & 0.9914 (0.0002) \\ 
\hline
\hline
\parbox[t]{2mm}{\multirow{7}{*}{\rotatebox[origin=c]{90}{Dataset \emph{SVHN-G}}}} & 
ADAM-MLE&0.9817&0.8446&0.8591    \\ 
\cline{2-5}
& ADAM-MAP&0.9812&0.8483&0.8619    \\ 
\cline{2-5}
& BBP&0.9826 &0.8567 &0.8701   \\ 
\cline{2-5}
& SGLD & 0.9806 (0.0002) & 0.8451 (0.0009)&0.8594 (0.0008)  \\ 
\cline{2-5}
& MCDropout & 0.9842 (0.0002) &0.8506 (0.0018) &0.8642 (0.0016) \\ 
\cline{2-5}
& SAE & 0.9837 (0.0022) & 0.8490 (0.0103)  &  0.8621 (0.0092) \\ 
\cline{2-5}
& {PMCnet-light}& \textbf{0.9880} (0)&\textbf{0.8742} (0.0003) & \textbf{0.8867} (0.0002) \\ 
\hline
\hline
\parbox[t]{2mm}{\multirow{7}{*}{\rotatebox[origin=c]{90}{Dataset \emph{CIFAR10}}}} & 
ADAM-MLE&0.9951&0.9149&0.9149    \\ 
\cline{2-5}
& ADAM-MAP&0.9954&0.9141&0.9142    \\ 
\cline{2-5}
& BBP&0.9968 &0.9310 &0.9310   \\ 
\cline{2-5}
& SGLD & 0.9881 (0.0010) & 0.8610 (0.0069)& 0.8600 (0.0070)  \\ 
\cline{2-5}
& MCDropout &0.9829 (0.0002)  & 0.8281 (0.0023) & 0.8287 (0.0023) \\ 
\cline{2-5}
& SAE & 0.9622 (0.0024)& 0.7414 (0.0080)  & 0.7419 (0.0080)  \\ 
\cline{2-5}
& {PMCnet-light}& \textbf{0.9972} (0)&\textbf{0.9364} (0.0002) & \textbf{0.9364} (0.0002) \\ 
\hline
\end{tabular}
}
\vspace{0.5cm}
\caption{\small Results for multi-class classification on datasets \emph{MNIST-C}, \emph{SVHN-G} and \emph{CIFAR10}, computed on test sets.}
\label{table:result6}
\end{table}

% \begin{table}[H]
% \hspace*{-1.2cm}
% \footnotesize
% \centering
% \begin{tabular}{|c|c|c|c|}
% \hline
%  \textbf{Method} & \textbf{AUC}& \textbf{F1 score} &\textbf{Accuracy} \\ 
% \hline
% ADAM-MLE&0.9998&0.9833&0.9834    \\ 
% \hline
% ADAM-MAP&0.9988&0.9547&0.9550    \\ 
% \hline
% BBP&0.9998 &0.9855 &0.9856   \\ 
% \hline
% SGLD & 0.9949 (0.0008) & 0.9115 (0.0135)&0.9126 (0.0132)  \\ 
% \hline
% MCDropout & 0.9998 (0) &0.9837 (0.0008) &0.9838 (0.0008) \\ 
% \hline
% SAE & 0.9998 (0) &0.9824 (0.0009) &\textbf{0.9998} (0.0003) \\ 
% \hline
% {PMCnet-light}& \textbf{1.0000} (0)&\textbf{0.9913} (0.0002) & 0.9914 (0.0002) \\ 
% \hline
% \end{tabular}
% \vspace{0.5cm}
% \caption{\small Results for multi-class classification on dataset \emph{MNIST-C}, computed on test set.}
% \label{table:result6}
% \end{table}

\begin{figure}[H]
\centering
\includegraphics[width = 14cm]{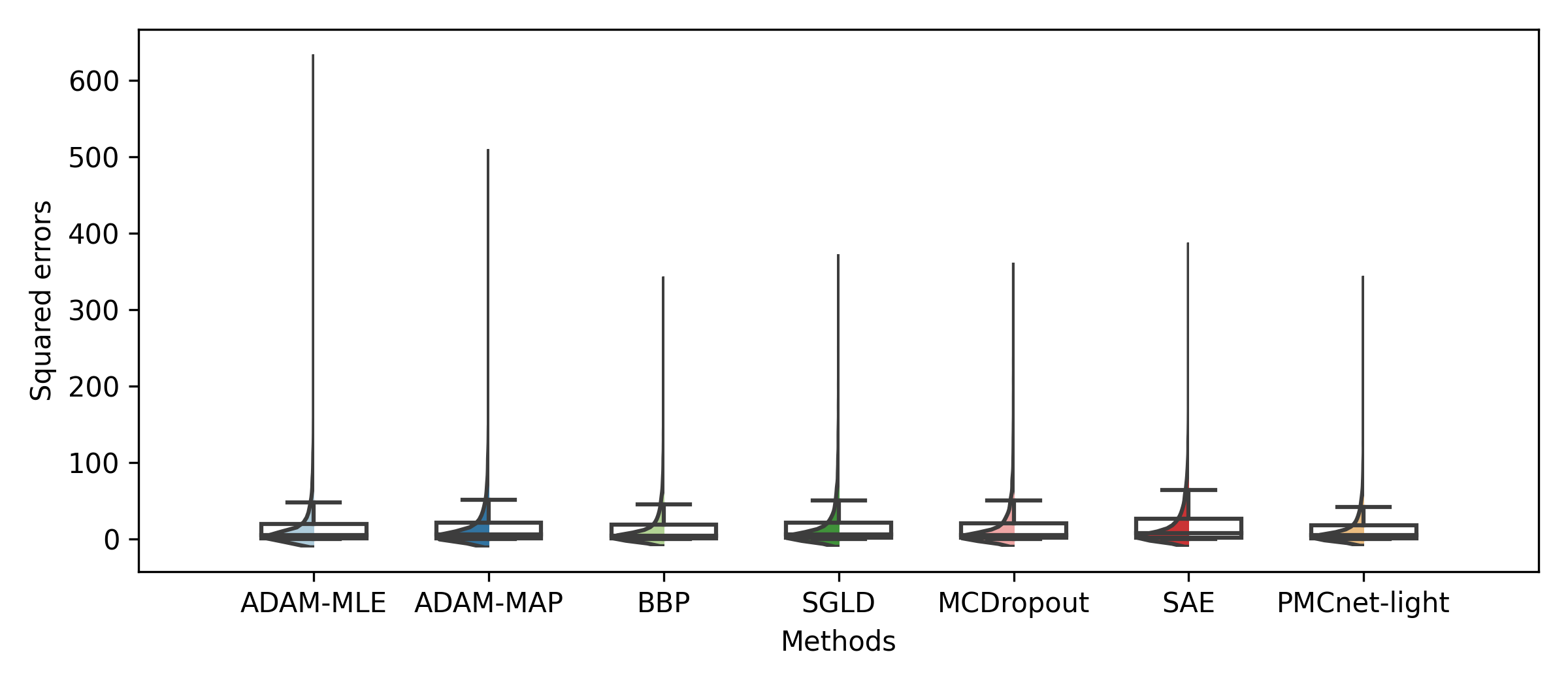}  
\caption{\footnotesize Box and violin plot of the squared errors of each sample in test set of dataset \emph{Protein}.}
\label{MSE:protein}
\end{figure}

\begin{figure}[H]
\centering
\includegraphics[width = 14cm]{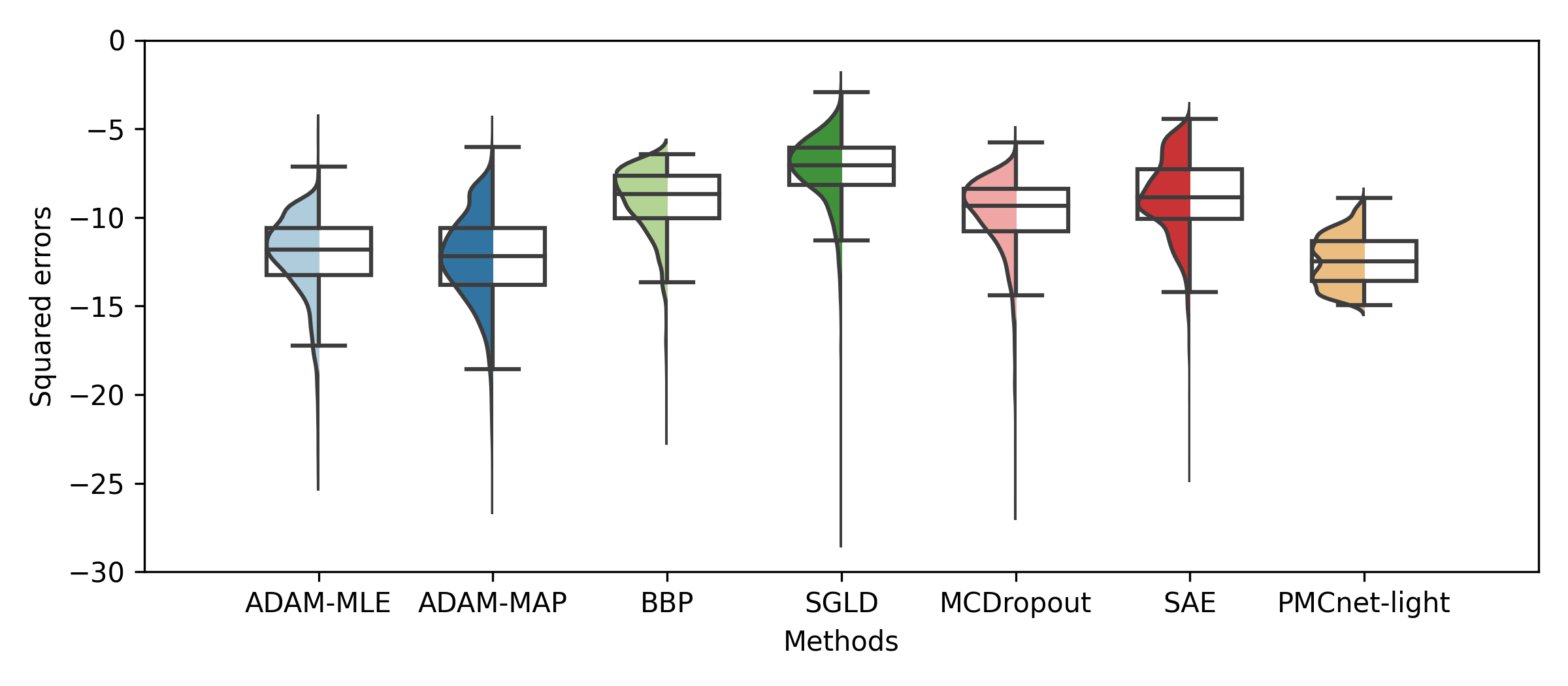}  
\caption{\footnotesize Box and violin plot of the squared errors of each sample in test set of dataset \emph{Naval}.}
\label{MSE:naval}
\end{figure}

\cblue{
We finally discuss the performance of the compared methods on the two regression datasets \emph{Protein} and \emph{Naval}.  The numerical results are summarized in Table~\ref{table:result7}, where we choose $T = 5$. The estimated output from our method has the smallest averaged MSE with rather small variability, for both datasets. We also display the box and violin plot of the square errors for each test examples, for both datasets, in Fig.~\ref{MSE:protein} and Fig.~\ref{MSE:naval}, respectively. Among all methods, {PMCnet-light} not only has the smallest MSE in average, but also shows minimal spreading of outliers, which shows again the good performance of our method.}

\begin{table}[H]
\footnotesize
\centering
\cblue{
\begin{tabular}{|c|c|c|}
\cline{2-3}
\multicolumn{1}{c|}{ } & \textbf{Method} & \textbf{MSE}\\ 
\hline
\parbox[t]{2mm}{\multirow{7}{*}{\rotatebox[origin=c]{90}{Dataset \emph{Protein}}}} & ADAM-MLE &17.8257  \\
\cline{2-3}
& ADAM-MAP &18.6500 \\
\cline{2-3}
& BBP &15.9396  \\
\cline{2-3}
& SGLD &17.4039 (0.2598)  \\
\cline{2-3}
& MCDropout & 17.3996 (0.1430) \\
\cline{2-3}
& SAE & 20.2291 (0.0297) \\
\cline{2-3}
& {PMCnet-light} &\textbf{15.2049} (0.0496)  \\
\hline
\hline
\parbox[t]{2mm}{\multirow{7}{*}{\rotatebox[origin=c]{90}{Dataset \emph{Naval}}}} & ADAM-MLE & $2.5171 \times 10^{-5}$ \\
\cline{2-3}
& ADAM-MAP &$4.6193 \times 10^{-5}$  \\
\cline{2-3}
& BBP & $3.1628\times 10^{-4}$   \\
\cline{2-3}
& SGLD & $1.6533 \times 10^{-3}$ (0.0002)  \\
\cline{2-3}
& MCDropout & $1.6158 \times 10^{-4}$ ($1.2333\times 10^{-5}$) \\
\cline{2-3}
& SAE & $5.5056\times 10^{-3}$ (0.0242) \\
\cline{2-3}
& {PMCnet-light} &\bm{$9.9032\times 10^{-6}$} ($2.8271\times 10^{-6}$)  \\
\hline
\end{tabular}
}
\vspace{0.5cm}
\caption{\small Results for regression problems on test sets of datasets \emph{Protein} and \emph{Naval}.}
\label{table:result7}
\end{table}

% \begin{table}[H]
% \footnotesize
% \centering
% \begin{tabular}{|c|c|}
% \hline
%  \textbf{Method} & \textbf{MSE}\\ 
% \hline
% ADAM-MLE &17.4054  \\
% \hline
% ADAM-MAP &18.6500 \\
% \hline
% BBP &15.9396  \\
% \hline
% SGLD &17.4039 (0.2598)  \\
% \hline
% MCDropout & 17.3996 (0.1430) \\
% \hline
% SAE & 20.2291 (0.0297) \\
% \hline
% {PMCnet-light} &\textbf{15.2049} (0.0496)  \\
% \hline
% \end{tabular}
% \vspace{0.5cm}
% \caption{\small Results for regression problem on test set of dataset \emph{Protein}.}
% \label{table:result7}
% \end{table}

\subsection{Computational complexity}

We finalize our experimental section by providing the computational time of different methods. We display the (total) training and test (averaged per sample) times in Table~\ref{table:time} for dataset \emph{Ionosphere}, and in Table~\ref{table:time_large} for dataset \emph{MNIST-C}. Note that the training times do not include the hyperparameter finetuning, necessary for all approaches. \cblue{We recall that our method, as well as SGLD, MCDropout and SAE, require a sampling procedure at test time in order to provide an approximation of the posterior distribution, which explains their higher computational cost, compared to ADAM and BBP that only provide pointwise output estimates.} In a nutshell, PMCnet approach is competitive in terms of computational time, both for training and test phases. In particular, the PMCnet-light variant has fast training, compared to the competitors achieving great performance.

% \begin{table}[H]
% \footnotesize
% \centering
% \begin{tabular}{|c||c|c|}
% \hline
% \textbf{Phase} &  \textbf{{PMCnet}}&  \textbf{{PMCnet-light}} \\
% \hline
% Training (s.) & 350 & 300 \\
% \hline
% Test (s.) & 2 & 2 \\ 
% \hline
% \end{tabular}
% \vspace{0.5cm}
% \caption{\small Computational time in seconds for each method during training phase (full time) and test phase (averaged time per sample) for the control dataset, using the same experimental scenario as in Table \ref{table:result_syn_diag}.}
% \label{table:time}
% \end{table}

\begin{table}[H]
\footnotesize
\centering
\begin{tabular}{|c||c|c|c|c|c|c|c|c|}
\hline
\textbf{Phase} & \textbf{ADAM-MLE}  & \textbf{ADAM-MAP} & \textbf{BBP}  & \textbf{SGLD} &\textbf{MCDropout}  &  \textbf{SAE} & \textbf{{PMCnet}}& \textbf{{PMCnet-light}} \\
\hline
Training (s.) & 108 & 152 & 195 &17   &10  &392 & 450  & 250\\
\hline
Test (ms.) & 1 & 1 & 1  & 43 &  89 &  112 & 239 & 239 \\
\hline
\end{tabular}
\vspace{0.5cm}
\caption{\small Computational time in seconds for each method during training phase (full time) and test phase (averaged time per sample) for dataset \emph{Ionosphere}, using the same experimental scenario as in Table \ref{table:result1}.}
\label{table:time}
\end{table}

\begin{table}[H]
\footnotesize
\centering
\begin{tabular}{|c||c|c|c|c|c|c|c|}
\hline
\textbf{Phase} & \textbf{ADAM-MLE}  & \textbf{ADAM-MAP} & \textbf{BBP}  & \textbf{SGLD} &\textbf{MCDropout}  & \textbf{SAE} &   \textbf{{PMCnet-light}}  \\
\hline
Training (s.) & 1760 & 1800  & 1035 &  560 & 680  &2100  & 1240 \\
\hline
Test (ms.) & 2 & 2 & 2 & 77  & 168 & 261  & 535\\
\hline
\end{tabular}
\vspace{0.5cm}
\caption{\small Computational time in seconds for each method during training phase (full time) and test phase (averaged time per sample) for dataset \emph{MNIST-C}, using same experimental scenario as in Table \ref{table:result6}.}
\label{table:time_large}
\end{table}

\section{Conclusion}
\label{sec:conclusion}
{This paper proposes a novel method for the Bayesian  inference in neural networks that relies on adaptive importance sampling. The method is able to characterize the posterior distribution of the estimated outputs and approximate the moments of such distribution. We also propose a light version of the method for handling large datasets and deep networks. This version avoids an excess of the required memory during the training process without impacting the performance of the algorithm. The numerical experiments illustrate the good performance of this new method on \cblue{several} datasets of classification and regression, using either shallow or deep networks. Our method is also competitive in terms of computational time on both training and test stages. \cblue{As a future work, we plan to adapt our approach to address more complex tasks such as face recognition and image segmentation.}}

\section{Declarations}

\cblue{\emph{Funding:} The work of Y. H. and E. C. is supported by the European Research Council Starting Grant MAJORIS ERC-2019-STG-850925. The work of V. E. is supported by the \emph{Agence Nationale de la Recherche} of France under PISCES (ANR-17-CE40-0031-01), the Leverhulme Research Fellowship (RF-2021-593), and by ARL/ARO under grants W911NF-20-1-0126 and W911NF-22-1-0235.
The work of J.-C. P. is supported by the ANR Chair in Artificial Intelligence BRIDGEABLE.
}

\cblue{\emph{Competing interests:} The authors have no relevant financial or non-financial interests
to disclose.}

\bibliography{refs}

\end{document}